\documentclass{article}

\usepackage[preprint]{neurips_data_2023}

\usepackage[utf8]{inputenc} 
\usepackage[T1]{fontenc}    
\usepackage{hyperref}       
\usepackage{url}            
\usepackage{booktabs}       
\usepackage{amsfonts}       
\usepackage{nicefrac}       
\usepackage{microtype}      
\usepackage[dvipsnames]{xcolor}
\usepackage{colortbl}
\usepackage{comment}
\usepackage{amsmath} 
\usepackage{adjustbox}
\usepackage{graphicx}
\usepackage{caption}
\usepackage{float}
\usepackage{cleveref}

\definecolor{Mycolor1}{HTML}{8FD9D6}
\definecolor{Mycolor2}{HTML}{BCEAA1}
\definecolor{Mycolor3}{HTML}{FDD598}

\title{HalluDial: A Large-Scale Benchmark for Automatic Dialogue-Level Hallucination Evaluation}

\author{%
Wen Luo$^{1,2}$\footnotemark[1] \quad Tianshu Shen$^3$ \quad Wei Li$^2$ \quad Guangyue Peng$^2$ \\ 
 \textbf{Richeng Xuan}$^1$ \quad \textbf{Houfeng Wang}$^2$\footnotemark[2] \quad \textbf{Xi Yang}$^1$\footnotemark[2] \\
$^1$Beijing Academy of Artificial Intelligence\\
$^2$National Key Laboratory for Multimedia Information Processing, Peking University\\
$^3$Wangxuan Institute of Computer Technology, Peking University}

\begin{document}

\maketitle

\renewcommand{\thefootnote}{\fnsymbol{footnote}}
\footnotetext[1]{Work done during an internship at Beijing Academy of Artificial Intelligence.}
\footnotetext[2]{Corresponding authors: \texttt{wanghf@pku.edu.cn, yangxi@baai.ac.cn}}

\begin{abstract}

Large Language Models (LLMs) have significantly advanced the field of Natural Language Processing (NLP), achieving remarkable performance across diverse tasks and enabling widespread real-world applications. However, LLMs are prone to hallucination, generating content that either conflicts with established knowledge or is unfaithful to the original sources. Existing hallucination benchmarks primarily focus on sentence- or passage-level hallucination detection, neglecting dialogue-level evaluation, hallucination localization, and rationale provision. They also predominantly target factuality hallucinations while underestimating faithfulness hallucinations, often relying on labor-intensive or non-specialized evaluators. To address these limitations, we propose HalluDial, the first comprehensive large-scale benchmark for automatic dialogue-level hallucination evaluation. HalluDial encompasses both spontaneous and induced hallucination scenarios, covering factuality and faithfulness hallucinations. The benchmark includes 4,094 dialogues with a total of 146,856 samples. Leveraging HalluDial, we conduct a comprehensive meta-evaluation of LLMs' hallucination evaluation capabilities in information-seeking dialogues and introduce a specialized judge language model, HalluJudge. The high data quality of HalluDial enables HalluJudge to achieve superior or competitive performance in hallucination evaluation, facilitating the automatic assessment of dialogue-level hallucinations in LLMs and providing valuable insights into this phenomenon. The dataset and the code are available at \url{https://github.com/FlagOpen/HalluDial}.

\end{abstract}

\section{Introduction}

Large Language Models (LLMs) have made considerable advancements in the field of Natural Language Processing (NLP). Their rapid evolution and widespread adoption have resulted in remarkable performance across diverse tasks \citep{bubeck2023sparks,zhao2023survey}, encompassing text generation, reading comprehension, and machine translation. This significant progress has facilitated their deployment in various real-world applications, serving millions of users globally.

Despite the outstanding capabilities exhibited by LLMs, they still suffer from the risk of hallucination. In the field of LLMs, hallucination refers to the phenomenon where the model generates content that either conflicts with established knowledge or is unfaithful to the original sources \citep{zhang2023siren,huang2023survey,li2024dawn}. This issue can result in producing inaccurate or misleading information, posing substantial challenges for the real-world deployments and applications of LLMs. Hence, the evaluation of hallucinations in LLMs has recently emerged as a critical task, leading to the proposal of numerous hallucination evaluation benchmarks \citep{manakul2023selfcheckgpt,chen2023beyond,zheng2023does,cheng2023evaluating,li2023halueval,liang2023uhgeval,yang2023new,feng2023factkb,min2023factscore}.

\begin{figure}
    \centering
    \includegraphics[width=0.95\textwidth]{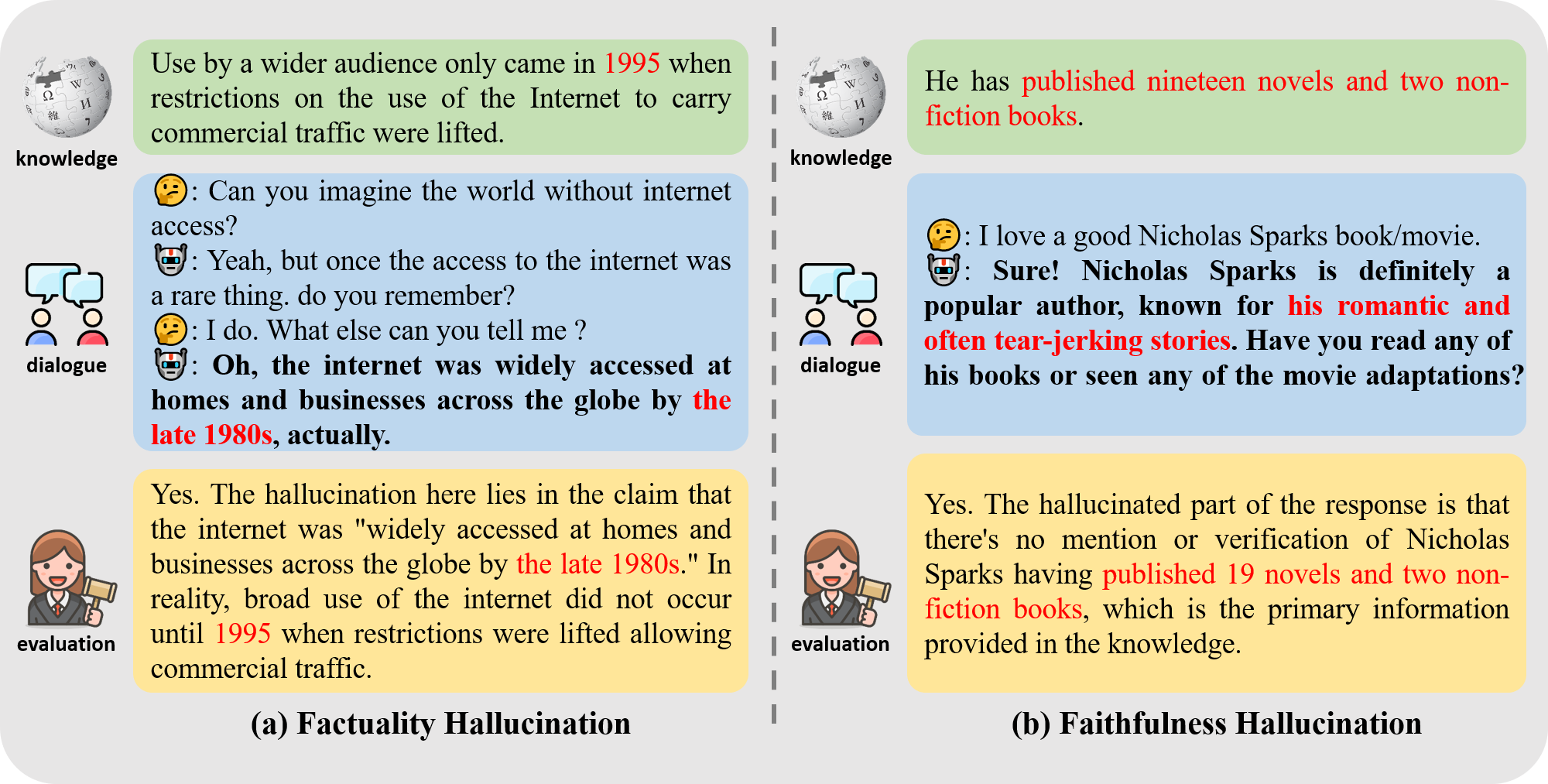}
    \caption{Samples from the HalluDial dataset, including knowledge, dialogue context, and hallucination evaluation results of the current response. Each evaluation result comprises hallucination detection, localization, and the corresponding rationale.}
    \label{fig:hallu_intro}
\end{figure}

Nevertheless, several limitations persist in these benchmarks:
(1) Absence of Dialogue-Level Hallucination Evaluation: Existing hallucination benchmarks primarily concentrate on sentence-level \citep{manakul2023selfcheckgpt,chen2023beyond,zheng2023does,cheng2023evaluating,li2023halueval} or passage-level \citep{liang2023uhgeval,yang2023new,feng2023factkb,min2023factscore} hallucinations. 
Although HaluEval \citep{li2023halueval} makes an effort to include dialogue-relevant topics, its main emphasis is still on the sentence and passage levels, and it only assesses hallucination detection capabilities.
Given the widespread deployment of LLMs in dialogue-based applications, the extent to which LLMs are prone to hallucinations at the dialogue level remains unclear.
(2) Neglect of Hallucination Localization and Rationale: The majority of existing evaluation work \citep{manakul2023selfcheckgpt,cheng2023evaluating,li2023halueval,yang2023new,min2023factscore} is merely centered on identifying the presence of hallucinations (i.e., hallucination detection), with little attention paid to the localization of hallucinations and the associated rationales, leading to the lack of interpretability to a certain extent.
(3) Predominant Focus on Factuality Hallucinations \citep{zhang2023siren,huang2023survey}: The focus of recent hallucination research in LLMs is mainly on factuality hallucinations, while the importance of faithfulness hallucinations, another crucial type of hallucination, is frequently underestimated.
(4) Reliance on Labor-Intensive Methods or Non-Specialized Evaluators: Current hallucination evaluations largely depend on crowd-sourced annotations or prompting LLMs that are not specifically designed for hallucination assessments, which are time-consuming and costly. Moreover, some research suggests that using API-based LLMs as evaluators can lead to issues such as inconsistency and irreproducibility \citep{wang2023pandalm}, while open-source LLMs still demonstrate limited performance as hallucination evaluators \citep{liang2023uhgeval}.

To address these limitations, we propose \textbf{HalluDial}, a comprehensive large-scale benchmark for automatic dialogue-level hallucination evaluation. HalluDial is derived from an information-seeking dialogue dataset \citep{dziri2022faithdial} and covers both factuality and faithfulness hallucinations. To understand the specific conditions under which LLMs naturally produce hallucinated content and simulate situations where these models intentionally generate hallucinated information, we meticulously design the spontaneous and induced hallucination scenarios, respectively. In the spontaneous hallucination scenario, we employ a two-step pipeline involving diverse dialogue sampling and automatic hallucination annotation. For the induced hallucination scenario, task-specific instructions are utilized to guide GPT-4 to directly generate hallucinated samples along with their explanations for each hallucination type. Combining these scenarios, HalluDial ultimately provides a large-scale benchmark comprising 146,856 samples, each containing results for hallucination detection, localization, and corresponding rationales (Figure \ref{fig:hallu_intro}).

By offering such a large and diverse dataset, HalluDial enables a comprehensive meta-evaluation of LLMs' capabilities in hallucination evaluation, including detection, localization, and rationale provision. Building on HalluDial, we introduce a specialized judge language model, HalluJudge. The high-quality data in HalluDial enables HalluJudge to achieve superior or competitive performance in hallucination evaluation, within HalluDial and in other generalization settings. Considering HalluJudge's outstanding performance, we subsequently employ it on HalluDial to automatically assess hallucinations in LLM-generated content during information-seeking dialogues, providing valuable insights into the nature and prevalence of hallucinations in LLMs.

Overall, the main contributions of this paper can be summarized as follows:

\begin{itemize}

\item We propose HalluDial, which, to the best of our knowledge, is the first large-scale dialogue-level hallucination benchmark.

\item We conduct a comprehensive meta-evaluation of LLMs' capabilities in hallucination evaluations and develop a hallucination judge language model named HalluJudge, which demonstrates superior or competitive capacity in HalluDial and other generalization settings.

\item We utilize HalluDial and HalluJudge to conduct an automatic evaluation of dialogue-level hallucination present in current LLMs.

\end{itemize}

\section{The HalluDial Benchmark}

To comprehensively understand the complexity of hallucination phenomena in LLMs during information-seeking conversations, we design two distinct scenarios: (1) the spontaneous hallucination scenario and (2) the induced hallucination scenario. 
The objective of the spontaneous hallucination scenario is to comprehend the frequency and specific conditions under which LLMs naturally produce hallucinated content during information-seeking dialogues. 
On the other hand, the induced hallucination scenario aims to simulate situations where disinformation, featuring hallucinated contents, is intentionally generated by these models. 
Data are constructed specifically for each scenario.

\subsection{Spontaneous Hallucination Scenario}
\label{sec: spontaneous_hallu}

In the spontaneous hallucination scenario, our data construction pipeline consists of two steps: (1) diverse dialogue sampling and (2) automatic hallucination annotation.

\paragraph{Diverse Dialogue Sampling}

In the diverse dialogue sampling, we begin by selecting a range of representative LLMs from a broad array of available options\footnote{\url{https://chat.lmsys.org}}. Given that different models can generate responses with varying degrees and aspects of hallucination, we carefully choose a suite of LLMs with varying parameter scales to ensure diversity in the generated responses. This suite includes Mistral-7B-Instruct-v0.2 \citep{jiang2023mistral}, vicuna-13B-v1.5 \citep{zheng2024judging}, vicuna-33B-v1.3 \citep{zheng2024judging}, Llama-2-70B-chat \citep{touvron2023llama}, and GPT-3.5-turbo.
Using the instruction templates presented in Table \ref{tab: diverese_dialogue_sampling}, we then instruct these selected models to engage in a natural information-seeking dialogue setting. Specifically, these models are tasked with generating a single response grounded in the provided external knowledge and the established dialogue history from FaithDial \citep{dziri2022faithdial}. Importantly, they are required to adhere to the length constraint and avoid introducing additional assumptions or automatically continuing the dialogue beyond the initial generated response. This setting aims to closely simulate the interactions that these LLMs might encounter in real-world applications and prepare consistent and coherent responses for annotation in the next stage.

\paragraph{Automatic Hallucination Annotation}

After collecting responses from a wide range of LLMs, the most direct approach to hallucination detection, localization, and providing explanation typically involves human annotation. However, this method is frequently costly and time-consuming, thereby lacking scalability. Given the robust capabilities of GPT-4 in following instructions and its proficiency in assessing hallucinations to a certain degree \citep{liang2023uhgeval}, we choose GPT-4 to perform automatic hallucination annotation on the collected natural responses. The key to our annotation approach lies in crafting effective instructions to guide GPT-4 systematically through the annotation process. As shown in Table \ref{tab: automatic_hallu_annotation}, the instruction prompt is carefully designed to encompass the following three components: (1) task definition, (2) example demonstration, and (3) annotation instructions. The task definition clearly outlines the input and the objective for GPT-4. To regulate the output format and enhance the model's comprehension of the task, we subsequently provide multiple examples for detailed guidance. To increase the reliability of the annotation results, we include the non-hallucinatory response for reference. Finally, the annotation instructions provide further directives on how to navigate the annotation process. Specifically, during the annotation process, based on the external knowledge, the dialogue history and the non-hallucinatory reference response, the assistant is required not only to identify any hallucinated information in the generated response, but also to refine its identification to the specific segment of the response that is inaccurate. Following this, the assistant should then provide reasonable and coherent justifications for its assessments. This detailed hallucination localization and rationales can enhance the interpretability of the evaluation results, contributing to a more profound understanding of hallucinations.

\subsection{Induced Hallucination Scenario}
\label{sec: induced_hallu}

The objective of the induced hallucination scenario is to emulate circumstances in which LLMs intentionally generate hallucinated information. In practical applications, LLMs will encounter a variety of requests, some of which could be malicious, aiming to deliberately generate hallucinated information to mislead others. Consequently, assessing how LLMs react when confronted with maliciously generated hallucinated information becomes crucial for their real-world deployment.
Within the framework of the induced hallucination scenario, for each type of hallucinations (i.e., factuality and faithfulness hallucinations), we devise type-specific instructions (Table \ref{tab: gen_fact_hallu}, \ref{tab: gen_faith_hallu}) to guide GPT-4 to directly produce hallucinated samples, based on the provided knowledge, the pre-existing dialogue history and the non-hallucinatory reference response. This scenario differs from the spontaneous hallucination scenario, which follows a sampling-then-annotating pipeline. To ensure data consistency and quality, in the induced hallucination scenario, GPT-4 is required to generate hallucinated samples while concurrently providing corresponding hallucination localization and rationales.

\subsection{Implementation Details}

The HalluDial dataset is constructed using GPT-4 version gpt-4-1106-preview. To ensure diversity, a temperature setting of 1.0 is applied during the diverse dialogue sampling and induced hallucination scenario. In contrast, a temperature of 0.0 is used during automatic hallucination annotation to minimize randomness and produce more deterministic outputs. Additionally, we set the frequency penalty to 0.0 and top-p to 1.0. The diverse dialogue sampling is conducted in a zero-shot manner to simulate the interactions that LLMs might encounter in real-world applications. For automatic hallucination annotation and the induced hallucination scenario, we provide three in-context examples to enhance the model's understanding of the task and ensure high data quality.

\subsection{Benchmark Statistics and Usage}

\begin{figure}
\centering
\begin{minipage}[!t]{0.30\textwidth}
\centering
\includegraphics[width=\textwidth]{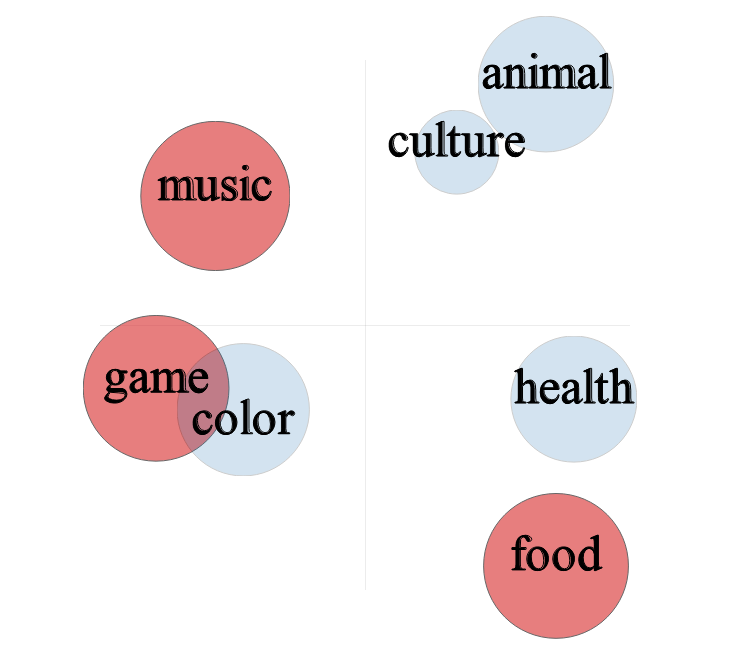}
\caption{Topic distributions in HalluDial dataset. The samples are categorized into seven topics, with the red circles highlighting the top topics.}
\label{fig:topic_halludial_1}
\end{minipage}
\hspace{0.05\textwidth}
\begin{minipage}[!t]{0.55\textwidth}
\centering
\captionof{table}{Statistics of HalluDial.}
\begin{tabular}{lc}
\toprule
\# Dialogues & 4,094\\
\# Turns & 18,357\\
\# Spontaneous Responses & 91,785\\
\quad \# Hallucinatory Responses & 21,706\\
\# Induced Responses &  55,071\\
\quad \# Factuality Hallucination Responses & 18,357\\
\quad \# Faithfulness Hallucination Responses & 18,357\\
\# Total Hallucinatory Responses & 58,420\\
\# Total Responses & 146,856\\
Avg. Turns per Dialogue & 4.48\\
Avg. Words per Model Utterance & 34.45\\
Avg. Words per Turn & 49.74\\
Avg. Words per Annotations & 25.26\\
\bottomrule
\end{tabular}
\label{tab:halluDialStats}
\end{minipage}
\end{figure}

As discussed in Section \ref{sec: spontaneous_hallu}, we annotate 91,785 natural responses from LLMs in the spontaneous hallucination scenario using a sampling-then-annotating pipeline. Of these, 21,706 are hallucinatory samples, and 70,079 are non-hallucinatory samples. In Section \ref{sec: induced_hallu}, we generate 18,357 hallucinated samples for each type of hallucinations. Combined with 18,357 non-hallucinatory samples from FaithDial \citep{dziri2022faithdial}, the HalluDial dataset ultimately comprises 146,856 data entries. Table \ref{tab:halluDialStats} provides the statistical details of HalluDial. In Figure \ref{fig:topic_halludial_1}, we further demonstrate the various topics covered in HalluDial, ranging from culture, music, and animals to food, games, and more.

Researchers can utilize the HalluDial dataset to investigate hallucinations of LLMs in several ways. First, by leveraging the hallucination detection labels in HalluDial, researchers can evaluate the hallucination detection capabilities of LLMs. Second, based on the hallucination localization and explanations provided in HalluDial, researchers can further assess the models' abilities to localize hallucinations and provide justifications. Third, researchers can train their own hallucination evaluators on HalluDial as needed and employ them to automatically assess whether the LLMs' generated contents contain hallucinations. Finally, researchers can analyze which topics or types of content pose greater challenges for hallucination detection or are more prone to generating hallucinations. 

\section{Evaluating Hallucination Detection, Localization and Explanation Capabilities}
\label{sec: meta-evaluation}

In this section, we utilize HalluDial to conduct a comprehensive meta-evaluation of LLMs, focusing on their abilities in hallucination evaluation, including hallucination detection, localization and providing explanations. Furthermore, to validate the quality of HalluDial, we have developed a specialized judge language model, HalluJudge, solely based on the training set of HalluDial\footnote{The implementation details are in Appendix \ref{sec: appendix_hallujudge_config}.}. Thorough experiments showcase HalluJudge’s superior performance and robust generalizability in handling tasks related to hallucination evaluations, revealing its potential as a competent evaluator for the hallucinations of other LLMs.

\subsection{Hallucination Detection}

We conduct hallucination recognition experiments\footnote{Please refer to Appendix \ref{sec: appendix_hallu_detect} for full results.} on the HalluDial test set to evaluate the ability of HalluJudge and other LLMs to detect hallucinations at the dialogue level. For the other LLMs, the evaluation instructions are detailed in Table \ref{tab: inst_hallu_detect}. The reported metrics include F1 scores for both hallucinatory and non-hallucinatory cases, overall accuracy (Acc.), average precision (P), average recall (R), and average F1 score (Macro F1) across both cases.
Based on the experimental results shown in Table \ref{tab:halludial_detect}, we observe the following findings: Firstly, HalluJudge demonstrates superior performance across all evaluation metrics, surpassing GPT-4 and GPT-4o, which indicates its excellent hallucination detection capability and the high data quality of HalluDial. Among the competing models, the GPT-4 family generally performs well, with GPT-4o-2024-05-13 achieving particularly strong results.
Furthermore, the hallucination detection capabilities of open-source LLMs and GPT-3.5-turbo are generally poor, suggesting that our proposed HalluDial benchmark poses a significant challenge to these models. Lastly, hallucination detection capabilities improve with increasing model parameter scale. From the perspective of the Macro F1 score, as the model parameter scale increases, its hallucination detection capability gradually improves. This indicates a significant impact of model parameter scale on performance. Possible reasons include that larger models typically possess more knowledge, stronger representational capabilities, and better adherence to instructions.

\begin{table}
\centering
\caption{Hallucination detection performance of HalluJudge and other LLMs on the HalluDial test set. Hallu.: Hallucinatory. Non-Hallu: Non-Hallucinatory.}
\begin{tabular}{lcccccc}
\toprule
 & Hallu. & Non-Hallu. & \multicolumn{4}{c}{Average}\\
 \cmidrule(lr){2-2} \cmidrule(lr){3-3} \cmidrule(lr){4-7}
 & F1 Score & F1 Score & Acc. & P & R & Macro F1\\
\midrule
Llama-2-7B-chat & 57.07 & 0.12 & 39.95 & 36.90 & 49.94 & 28.59\\
Llama-2-13B-chat & 57.41 & 11.64 & 42.52 & 57.52 & 51.52 & 34.53\\
Llama-2-70B-chat & 59.37 & 60.93 & 60.16 & 62.09 & 62.26 & 60.15\\
vicuna-7B-v1.5 & 19.41 & 76.44 & 63.54 & 73.00  & 54.77 & 47.92\\
vicuna-13B-v1.5 & 31.25 & 78.30 & 67.01 & 79.22 & 58.96 & 54.77\\
vicuna-33B-v1.3 & 56.26 & 61.63 & 59.12 & 59.86 & 60.23 & 58.94\\
GPT-3.5-turbo & 54.32 & 81.43 & 73.60 & 79.29 & 67.87 & 67.88\\
GPT-4-1106-preview & 67.75 & 85.02 & 79.54 & 83.75 & 75.24 & 76.39\\
GPT-4-0125-preview & 74.15 & 86.18 & 81.99 & 83.43 & 79.09 & 80.16\\
GPT-4o-2024-05-13 & 72.66 & 85.01 & 80.64 & 81.42 & 77.92 & 78.84\\
HalluJudge & \textbf{80.07} & \textbf{89.77} & \textbf{86.48} & \textbf{89.89} & \textbf{83.38} & \textbf{84.92}\\
\bottomrule
\end{tabular}
\label{tab:halludial_detect}
\end{table}

\subsection{Hallucination Localization and Explanation}

In this section, we evaluate LLMs' capability in hallucination localization and explanation, employing a random selection of 3200 test instances sourced from the HalluDial test set. For the other LLMs, the instruction with three in-context examples is present in Table \ref{tab: inst_hallu_rationales}. We first utilize established automatic evaluation metrics, including ROUGE\_L, BLEU-4, and BERTScore \citep{zhang2019bertscore}, to measure the similarity between the generated hallucination localization and explanation by the models against the reference within the HalluDial dataset.
To provide a more holistic assessment of the models' performance in hallucination localization and explanation, human annotations are employed. Specifically, three annotators are required to evaluate the reasonableness and accuracy of the localization and explanation produced by GPT-3.5-turbo, GPT-4-1106-preview, and HalluJudge. We use Cohen's Kappa ($\kappa$) to measure the inter-annotator agreement (IAA) and obtain an average $\kappa=0.902(0.80\leq\kappa\leq1.00)$, indicating a perfect consistency among annotators.
The results presented in Table \ref{tab:halludial_rationale} show that HalluJudge notably outperforms other models across multiple metrics, including ROUGE\_L, BERTScore, and human evaluations. This suggests not only its effectiveness in identifying the presence of hallucinations but also its ability to accurately localize them and provide coherent rationales.

\begin{table}
\caption{The performance of LLMs in hallucination localization and providing rationales. Human denotes the results of human annotations.}
\centering
\begin{tabular}{lcccc}
\toprule
 & ROUGE\_L & BLEU-4 & BERTScore & Human\\
\midrule
Llama-2-7B-chat & 6.42 & 2.36 & 35.69 & - \\
Llama-2-13B-chat & 11.53 & 5.55 & 40.79 & - \\
Llama-2-70B-chat & 14.80 & 8.03 & 44.66 & - \\
vicuna-7B-v1.5 & 45.26 & 13.56 & 63.60 & - \\
vicuna-13B-v1.5 & 46.12 & 16.48 & 65.00 & - \\
vicuna-33B-v1.3 & 21.09 & 13.10 & 49.80 & - \\
GPT-3.5-turbo & 52.58 & 29.88 & 69.49 & 83.41\\
GPT-4-1106-preview & 72.38 & \textbf{39.97} & 83.09 & 89.41 \\
GPT-4-0125-preview & 73.01 & 24.36 & 83.09 & - \\
GPT-4o-2024-05-13 & 72.00 & 37.33 & 82.92 & - \\
HalluJudge & \textbf{77.38} & 32.06 & \textbf{86.38} & \textbf{93.65}\\
\bottomrule
\end{tabular}
\label{tab:halludial_rationale}
\end{table}

\subsection{Generalizability of HalluJudge}

To further ensure the high data quality of HalluDial, we incorporate the HaluEval dataset \citep{li2023halueval} to evaluate the out-of-distribution and cross-domain generalizability of HalluJudge. The HaluEval dataset is a hallucination detection benchmark that includes three sub-tasks: dialogue, question answering (QA), and summarization. Each sub-task comprises 10,000 hallucinatory samples and 10,000 non-hallucinatory samples. To assess the out-of-distribution generalizability of HalluJudge, we utilize the dialogue task from HaluEval. For evaluating cross-domain generalizability, we employ the QA and summarization tasks. In our experimental setup, we train the model on the HalluDial training set and test it on the target datasets. The other models used for comparison possess comparable overall costs.
The results presented in Tables \ref{tab:ood_general} demonstrate that HalluJudge exhibits superior performance in out-of-distribution settings, outperforming GPT-3.5-turbo. While in cross-domain settings, as shown in Table \ref{tab:cross_domain_general}, HalluJudge achieves competitive performances for the QA task and surpasses GPT-3.5-turbo in summarization in terms of accuracy. These findings underscore the robust generalizability of HalluJudge and the high quality of the HalluDial dataset. Furthermore, a comparison of GPT-3.5-turbo's performance on both HalluDial and the dialogue subset of HaluEval reveals that HalluDial represents a more challenging benchmark for hallucination detection in existing LLMs.

\begin{table}
\caption{Out-of-distribution generalizability of HalluJudge on the HaluEval dataset.}
\centering
\begin{tabular}{lcccccc}
\toprule
 & Hallu. & Non-Hallu.  & \multicolumn{4}{c}{Average}\\
 \cmidrule(lr){2-2} \cmidrule(lr){3-3} \cmidrule(lr){4-7}
 & F1 Score & F1 Score & Acc. & P & R & Macro F1\\
\midrule
Llama-2-7B-chat & 66.64 & 0.02 & 49.97 & 32.13 & 49.97 & 33.33\\
Llama-2-13B-chat & 69.83 & 2.12 & 53.87 & 72.15 & 50.49 & 35.97\\
Llama-2-70B-chat & 68.09 & 14.02 & 53.45 & 71.79 & 53.45 & 41.06\\
vicuna-7B-v1.5 & 43.58 & 71.29 & 61.94 & 70.74 & 61.94 & 57.43\\
vicuna-13B-v1.5 & 65.76 & 76.59 & 72.19 & 75.84 & 72.20 & 71.18\\
vicuna-33B-v1.3 & 69.37 & 37.42 & 58.88 & 66.75 & 58.88 & 53.40\\
GPT-3.5-turbo & 78.37 & 73.80 & 76.30 & 77.30 & 76.30 & 76.08\\
HalluJudge & \textbf{82.23} & \textbf{83.04} & \textbf{82.64} & \textbf{82.72} & \textbf{82.64} & \textbf{82.64}\\
\bottomrule
\end{tabular}
\label{tab:ood_general}
\end{table}

\begin{table}
\caption{Cross-domain generalizability of HalluJudge on the HaluEval dataset. $\dagger$ denotes the results are from \citet{li2023halueval}.}
\centering
\begin{adjustbox}{width=\textwidth}
\begin{tabular}{lcccccccc}
\toprule
 & \multicolumn{4}{c}{QA} & \multicolumn{4}{c}{Summarization}\\
\cmidrule(lr){2-5} \cmidrule(lr){6-9}
 & Acc. & Avg.P & Avg.R & Macro F1 & Acc. & Avg.P & Avg.R & Macro F1\\
\midrule
Llama-2-7B-chat & 49.94 & 28.56 & 49.94 & 33.32 & 49.46 & 29.74 & 49.96 & 33.10\\
Llama-2-13B-chat & 61.65 & 73.97 & 60.78 & 55.07 & 54.37 & 54.58 & 54.32 & 53.68\\
Llama-2-70B-chat & 68.48 & 68.86 & 68.48 & 68.33 & 49.45 & 42.26 & 49.45 & 34.16\\
vicuna-7B-v1.5 & 54.73 & 71.64 & 54.73 & 43.73 & 52.84 & 53.61 & 52.84 & 50.16\\
vicuna-13B-v1.5 & 65.34 & 77.62 & 65.34 & 61.01 & 53.01 & 72.09 & 53.01 & 40.06\\
vicuna-33B-v1.3 & 60.39 & 61.34 & 60.39 & 59.54 & 51.89 & 52.03 & 51.88 & 50.94\\
GPT-3.5-turbo & \textbf{74.47} & 81.54 & \textbf{74.97} & \textbf{72.95} & 58.53$\dagger$ & - & - & - \\
HalluJudge & 71.34 & \textbf{81.78} & 71.34 & 68.78 & \textbf{59.54} & \textbf{76.74} & \textbf{59.54} & \textbf{51.79}\\
\bottomrule
\end{tabular}
\end{adjustbox}
\label{tab:cross_domain_general}
\end{table}

\section{Evaluating the Hallucinations of LLMs in Information-Seeking Dialogues}
\label{sec: evaluate_hallu_rate}

As discussed in Section \ref{sec: meta-evaluation}, the high data quality of HalluDial enables HalluJudge to effectively detect and localize hallucinations, while also providing reasonable and coherent rationales. Consequently, we leverage HalluJudge to automatically evaluate hallucinations in other LLMs during information-seeking dialogues. Specifically, we utilize the templates outlined in Table \ref{tab: diverese_dialogue_sampling} to guide LLMs in generating natural responses based on the given dialogue history and the associated external knowledge from the HalluDial dataset. Following this, HalluJudge is used to assess whether the generated responses contain any hallucinations.

\subsection{Main Results}

Table \ref{tab: hallu_rate} presents the hallucination rates\footnote{The full results are listed in Appendix \ref{sec: appendix_hallu_rate}.} of various LLMs on HalluDial\footnote{The temperature is set to 1.0.}. The evaluation encompasses two scenarios (i.e., Gold and Self), each examined across two dimensions (i.e., turn and dialogue). The Gold scenario refers to instances where the previous dialogue history is populated with non-hallucinatory reference responses, while the Self scenario involves dialogue histories filled with responses generated by the evaluated model itself. The turn dimension represents the hallucination rate at the turn level, while the dialogue dimension represents the hallucination rate at the dialogue level (i.e., if any turn within a dialogue contains hallucinations, the entire dialogue is considered hallucinatory).
As shown in Table \ref{tab: hallu_rate}, it is evident that most LLMs maintain relatively low hallucination rates in the context of information-seeking dialogues. This finding suggests that current LLMs can generally produce relatively high-quality responses when provided with accurate and appropriate external knowledge for reference. However, the hallucination rates are not uniformly consistent between the Self and Gold scenarios. Most models exhibit lower hallucination rates in the Self scenario compared to the Gold scenario. This implies that in knowledge-grounded settings, using the model's own generated responses rather than non-hallucinatory reference responses in the history may help reduce hallucinations. Conversely, this discrepancy might also be attributable to a loss of dialogue coherence when the dialogue history is filled with the model's own generated responses.

\begin{table}
\caption{Hallucination rates of LLMs.}
\centering
\begin{tabular}{lcccc}
\toprule
 & \multicolumn{2}{c}{Gold.} & \multicolumn{2}{c}{Self.}\\
\cmidrule(lr){2-3} \cmidrule(lr){4-5}
 & turn. & dialogue. & turn. & dialogue.\\
\midrule
Llama-2-7B-chat	& 10.87 & 36.96 & 10.62 & 36.68\\
Llama-2-13B-chat & 10.10 & 35.11 & 9.09 & 32.39\\
Llama-2-70B-chat & 9.81 & 34.35 & 8.52 & 31.14\\
vicuna-7B-v1.5 & 4.39 & 17.68 & 4.31 & 17.39\\
vicuna-13B-v1.5 & 7.01 & 26.32 & 6.40 & 24.81\\
vicuna-33B-v1.3 & 6.10 & 23.97 & 5.08 & 20.14\\
Mistral-7B-Instruct-v0.2 & 3.14 & 12.91 & 2.74 & 11.51\\
GPT-3.5-turbo & 3.79 & 13.43 & 3.17 & 12.62\\
GPT-4-1106-preview & 5.52 & 22.72 & 5.52 & 22.33\\
GPT-4-0125-preview & 8.85 & 32.90 & 6.70 & 26.01\\
GPT-4o-2024-05-13 & 2.77 & 11.70 & 2.53 & 10.75\\
\bottomrule
\end{tabular}
\label{tab: hallu_rate}
\end{table}

\subsection{Impact of Temperature on Hallucinations}

\begin{figure*}[!htb]
  \centering
  \includegraphics[width=0.95\textwidth]{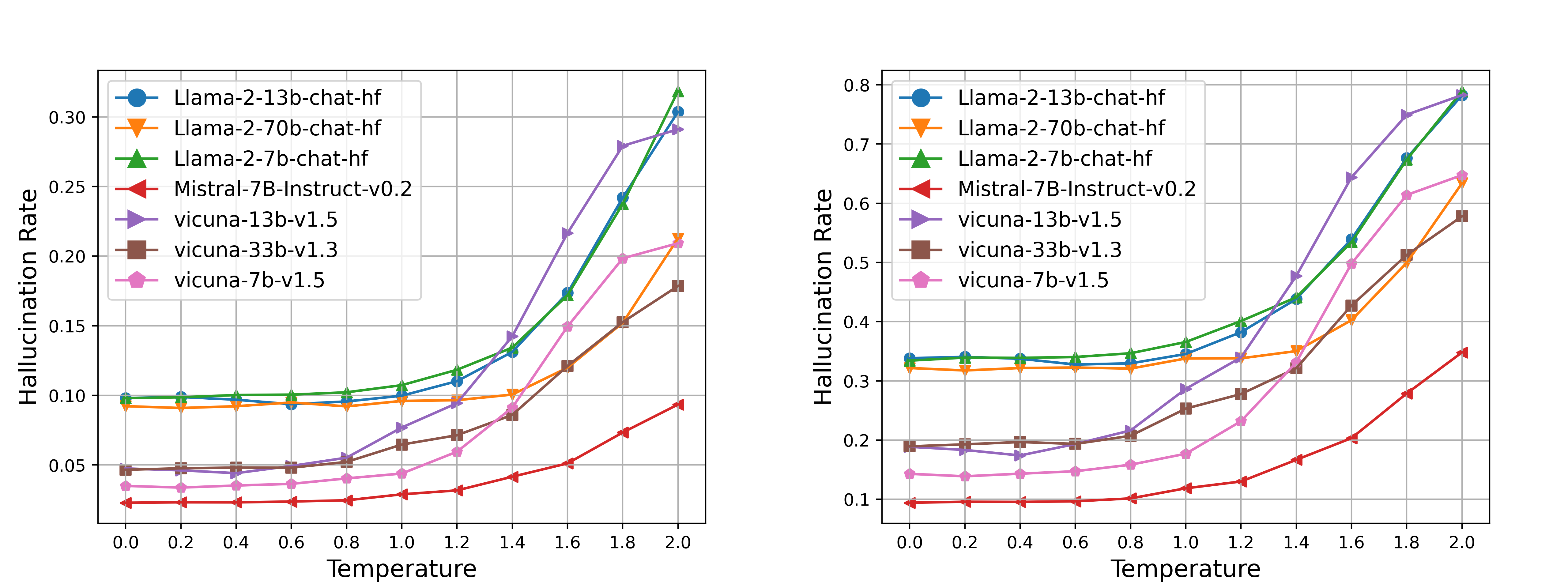}
  \caption{Impact of temperature on hallucination rate. Left: turn level. Right: dialogue level.}
\label{fig: hallu_temp}
\end{figure*}

In this section, we investigate the relationship between sampling temperature and the hallucination tendency of LLMs. As shown in Figure \ref{fig: hallu_temp}, we initially observe a consistent trend where the hallucination rate increases with rising temperature, aligning with prior research findings. Specifically, we note that as temperature increases, both turn-level and dialogue-level hallucination rates demonstrate a period of relative stability, followed by a sudden and significant rise.
Specifically, at lower sampling temperatures, the hallucination rate remains relatively low and stable. However, as the temperature continues to climb, there is a marked increase in the hallucination rate. 
Temperature is a crucial factor influencing the randomness in the generation process of LLMs. Lower temperatures tend to produce more stable and consistent results, while higher temperatures can enhance the randomness and diversity of generated outputs. These observations highlight the dual effects of temperature on model generation: while a higher temperature can increase the diversity of generated content, it also raises the incidence of hallucinations. Therefore, in practical applications, it is essential to select an appropriate temperature value to balance the diversity of generated content and the degree of hallucinations, particularly in scenarios requiring accurate and factual responses.

\section{Related Work}

\paragraph{Hallucinations Benchmarks}

Hallucination refers to the phenomenon where a model generates content that either conflicts with established knowledge or is unfaithful to the original sources. To address this issue, numerous benchmarks have been developed. Some studies assess hallucinations by integrating external knowledge sources \citep{guo2022survey,zhong2020reasoning,min2023factscore}. Other research employs question-answering mechanisms to test the factual accuracy of model responses \citep{zhao2023knowing,chen2023beyond,huang2023look,zheng2023does,yin2023large,cheng2023evaluating}.
TruthfulQA \citep{lin2022truthfulqa} is a benchmark designed to assess imitative falsehoods resulting from misleading pretraining data. HalluQA \citep{cheng2023evaluating} contains 450 adversarial questions to evaluate the hallucinations of Chinese LLMs. For automatic evaluation, SelfCheckGPT \citep{manakul2023selfcheckgpt} leverages the consistency of LLMs' outputs to design a sentence-level hallucination assessment metric. \citet{yang2023new} proposes a passage-level hallucination detection benchmark. HaluEval \citep{li2023halueval} is a hallucination detection benchmark containing GPT-3.5-turbo-generated and human-annotated hallucinated samples. UHGEval \citep{liang2023uhgeval} focuses on assessing hallucinations of Chinese LLMs in unconstrained settings.
Despite the progress made by existing research, several limitations remain. Notably, there is a lack of dialogue-level hallucination evaluation, with current benchmarks focusing mainly on the sentence level \citep{zhao2023knowing,chen2023beyond,huang2023look,zheng2023does,yin2023large,cheng2023evaluating,li2023halueval} and passage level \citep{zhong2020reasoning,min2023factscore,manakul2023selfcheckgpt,liang2023uhgeval}. Since dialogue is one of the most common applications of LLMs, a comprehensive evaluation of dialogue-level hallucination remains notably absent. Additionally, existing approaches often overlook hallucination localization and associated rationales, neglect faithfulness hallucinations, and rely on labor-intensive methods or non-specialized evaluators. In this study, we aim to address these limitations and introduce HalluDial, a large-scale benchmark for automatic dialogue-level hallucination evaluation.

\paragraph{Evaluations with LLMs}

As the development of Large Language Models (LLMs) continues to progress, an increasing volume of research has begun to investigate their potential for substituting human evaluators \citep{cheng2023improving,zheng2024judging,li2023alpacaeval,fu2023gptscore}. One notable framework is GPTScore \citep{fu2023gptscore}, which leverages LLMs to evaluate generated text. Moreover, \citet{zheng2024judging} delves into the application of LLMs as evaluators for other conversational agents, demonstrating that the assessments made by GPT-4 align with human preferences approximately 80\% of the time.
Current efforts in hallucination evaluation often utilize either open-source LLMs \citep{manakul2023selfcheckgpt,li2024attributionbench,yue2023automatic} or API-based LLMs \citep{min2023factscore,yang2023new,li2023halueval,liang2023uhgeval}. However, it is important to note that the use of API-based LLMs as evaluators has been associated with issues related to inconsistency and irreproducibility \citep{wang2023pandalm}. Furthermore, open-source LLMs have yet to yield satisfactory results when used as hallucination evaluators \citep{liang2023uhgeval}. Therefore, building on HalluDial, we further develop a specialized judge language model for automatic evaluation.

\section{Conclusion}

In this paper, we introduced HalluDial, the first large-scale benchmark for dialogue-level hallucination evaluation, addressing critical gaps in existing benchmarks. HalluDial encompasses both spontaneous and induced hallucination scenarios, covering factuality and faithfulness hallucinations. It provides a comprehensive dataset of 146,856 samples with detailed results for hallucination detection, localization, and rationale provision, which enables a thorough meta-evaluation of LLMs' capabilities in hallucination evaluation. Furthermore, the high data quality of HalluDial supports the development of HalluJudge, a specialized hallucination judge language model, and facilitates the automatic assessment of dialogue-level hallucinations in LLMs. In the future, we are committed to further enhancing the reliability of automatic hallucination evaluation by developing increasingly robust and generalized versions of HalluJudge.

\bibliographystyle{bib_style}
\bibliography{bib}

\begin{thebibliography}{32}
\providecommand{\natexlab}[1]{#1}
\providecommand{\url}[1]{\texttt{#1}}
\expandafter\ifx\csname urlstyle\endcsname\relax
  \providecommand{\doi}[1]{doi: #1}\else
  \providecommand{\doi}{doi: \begingroup \urlstyle{rm}\Url}\fi

\bibitem[Bubeck et~al.(2023)Bubeck, Chandrasekaran, Eldan, Gehrke, Horvitz, Kamar, Lee, Lee, Li, Lundberg, et~al.]{bubeck2023sparks}
S{\'e}bastien Bubeck, Varun Chandrasekaran, Ronen Eldan, Johannes Gehrke, Eric Horvitz, Ece Kamar, Peter Lee, Yin~Tat Lee, Yuanzhi Li, Scott Lundberg, et~al.
\newblock Sparks of artificial general intelligence: Early experiments with gpt-4.
\newblock \emph{arXiv preprint arXiv:2303.12712}, 2023.

\bibitem[Chen et~al.(2023)Chen, Deng, Bian, Qin, Wu, Chua, and Wong]{chen2023beyond}
Liang Chen, Yang Deng, Yatao Bian, Zeyu Qin, Bingzhe Wu, Tat-Seng Chua, and Kam-Fai Wong.
\newblock Beyond factuality: A comprehensive evaluation of large language models as knowledge generators.
\newblock In \emph{Proceedings of the 2023 Conference on Empirical Methods in Natural Language Processing}, pp.\  6325--6341, 2023.

\bibitem[Cheng et~al.(2023{\natexlab{a}})Cheng, Sun, Zhang, Wang, Liu, Zhang, He, Huang, Yin, Chen, et~al.]{cheng2023evaluating}
Qinyuan Cheng, Tianxiang Sun, Wenwei Zhang, Siyin Wang, Xiangyang Liu, Mozhi Zhang, Junliang He, Mianqiu Huang, Zhangyue Yin, Kai Chen, et~al.
\newblock Evaluating hallucinations in chinese large language models.
\newblock \emph{arXiv preprint arXiv:2310.03368}, 2023{\natexlab{a}}.

\bibitem[Cheng et~al.(2023{\natexlab{b}})Cheng, Yang, Sun, Li, and Qiu]{cheng2023improving}
Qinyuan Cheng, Xiaogui Yang, Tianxiang Sun, Linyang Li, and Xipeng Qiu.
\newblock Improving contrastive learning of sentence embeddings from ai feedback.
\newblock In \emph{Findings of the Association for Computational Linguistics: ACL 2023}, pp.\  11122--11138, 2023{\natexlab{b}}.

\bibitem[Dziri et~al.(2022)Dziri, Kamalloo, Milton, Zaiane, Yu, Ponti, and Reddy]{dziri2022faithdial}
Nouha Dziri, Ehsan Kamalloo, Sivan Milton, Osmar Zaiane, Mo~Yu, Edoardo~M Ponti, and Siva Reddy.
\newblock Faithdial: A faithful benchmark for information-seeking dialogue.
\newblock \emph{Transactions of the Association for Computational Linguistics}, 10:\penalty0 1473--1490, 2022.

\bibitem[Feng et~al.(2023)Feng, Balachandran, Bai, and Tsvetkov]{feng2023factkb}
Shangbin Feng, Vidhisha Balachandran, Yuyang Bai, and Yulia Tsvetkov.
\newblock Factkb: Generalizable factuality evaluation using language models enhanced with factual knowledge.
\newblock In \emph{Proceedings of the 2023 Conference on Empirical Methods in Natural Language Processing}, pp.\  933--952, 2023.

\bibitem[Fu et~al.(2023)Fu, Ng, Jiang, and Liu]{fu2023gptscore}
Jinlan Fu, See-Kiong Ng, Zhengbao Jiang, and Pengfei Liu.
\newblock Gptscore: Evaluate as you desire.
\newblock \emph{arXiv preprint arXiv:2302.04166}, 2023.

\bibitem[Guo et~al.(2022)Guo, Schlichtkrull, and Vlachos]{guo2022survey}
Zhijiang Guo, Michael Schlichtkrull, and Andreas Vlachos.
\newblock A survey on automated fact-checking.
\newblock \emph{Transactions of the Association for Computational Linguistics}, 10:\penalty0 178--206, 2022.

\bibitem[Huang et~al.(2023{\natexlab{a}})Huang, Yu, Ma, Zhong, Feng, Wang, Chen, Peng, Feng, Qin, et~al.]{huang2023survey}
Lei Huang, Weijiang Yu, Weitao Ma, Weihong Zhong, Zhangyin Feng, Haotian Wang, Qianglong Chen, Weihua Peng, Xiaocheng Feng, Bing Qin, et~al.
\newblock A survey on hallucination in large language models: Principles, taxonomy, challenges, and open questions.
\newblock \emph{arXiv preprint arXiv:2311.05232}, 2023{\natexlab{a}}.

\bibitem[Huang et~al.(2023{\natexlab{b}})Huang, Song, Wang, Chen, and Ma]{huang2023look}
Yuheng Huang, Jiayang Song, Zhijie Wang, Huaming Chen, and Lei Ma.
\newblock Look before you leap: An exploratory study of uncertainty measurement for large language models.
\newblock \emph{arXiv preprint arXiv:2307.10236}, 2023{\natexlab{b}}.

\bibitem[Jiang et~al.(2023)Jiang, Sablayrolles, Mensch, Bamford, Chaplot, de~las Casas, Bressand, Lengyel, Lample, Saulnier, Lavaud, Lachaux, Stock, Scao, Lavril, Wang, Lacroix, and Sayed]{jiang2023mistral}
Albert~Q. Jiang, Alexandre Sablayrolles, Arthur Mensch, Chris Bamford, Devendra~Singh Chaplot, Diego de~las Casas, Florian Bressand, Gianna Lengyel, Guillaume Lample, Lucile Saulnier, Lélio~Renard Lavaud, Marie-Anne Lachaux, Pierre Stock, Teven~Le Scao, Thibaut Lavril, Thomas Wang, Timothée Lacroix, and William~El Sayed.
\newblock Mistral 7b, 2023.

\bibitem[Li et~al.(2023{\natexlab{a}})Li, Cheng, Zhao, Nie, and Wen]{li2023halueval}
Junyi Li, Xiaoxue Cheng, Wayne~Xin Zhao, Jian-Yun Nie, and Ji-Rong Wen.
\newblock Halueval: A large-scale hallucination evaluation benchmark for large language models.
\newblock In \emph{Proceedings of the 2023 Conference on Empirical Methods in Natural Language Processing}, pp.\  6449--6464, 2023{\natexlab{a}}.

\bibitem[Li et~al.(2024{\natexlab{a}})Li, Chen, Ren, Cheng, Zhao, Nie, and Wen]{li2024dawn}
Junyi Li, Jie Chen, Ruiyang Ren, Xiaoxue Cheng, Wayne~Xin Zhao, Jian-Yun Nie, and Ji-Rong Wen.
\newblock The dawn after the dark: An empirical study on factuality hallucination in large language models.
\newblock \emph{arXiv preprint arXiv:2401.03205}, 2024{\natexlab{a}}.

\bibitem[Li et~al.(2023{\natexlab{b}})Li, Zhang, Dubois, Taori, Gulrajani, Guestrin, Liang, and Hashimoto]{li2023alpacaeval}
Xuechen Li, Tianyi Zhang, Yann Dubois, Rohan Taori, Ishaan Gulrajani, Carlos Guestrin, Percy Liang, and Tatsunori~B Hashimoto.
\newblock Alpacaeval: An automatic evaluator of instruction-following models, 2023{\natexlab{b}}.

\bibitem[Li et~al.(2024{\natexlab{b}})Li, Yue, Liao, and Sun]{li2024attributionbench}
Yifei Li, Xiang Yue, Zeyi Liao, and Huan Sun.
\newblock Attributionbench: How hard is automatic attribution evaluation?, 2024{\natexlab{b}}.

\bibitem[Liang et~al.(2023)Liang, Song, Niu, Li, Xiong, Tang, Wy, He, Cheng, Wang, et~al.]{liang2023uhgeval}
Xun Liang, Shichao Song, Simin Niu, Zhiyu Li, Feiyu Xiong, Bo~Tang, Zhaohui Wy, Dawei He, Peng Cheng, Zhonghao Wang, et~al.
\newblock Uhgeval: Benchmarking the hallucination of chinese large language models via unconstrained generation.
\newblock \emph{arXiv preprint arXiv:2311.15296}, 2023.

\bibitem[Lin et~al.(2022)Lin, Hilton, and Evans]{lin2022truthfulqa}
Stephanie Lin, Jacob Hilton, and Owain Evans.
\newblock Truthfulqa: Measuring how models mimic human falsehoods.
\newblock In \emph{Proceedings of the 60th Annual Meeting of the Association for Computational Linguistics (Volume 1: Long Papers)}, pp.\  3214--3252, 2022.

\bibitem[Loshchilov \& Hutter(2018)Loshchilov and Hutter]{loshchilov2018decoupled}
Ilya Loshchilov and Frank Hutter.
\newblock Decoupled weight decay regularization.
\newblock In \emph{International Conference on Learning Representations}, 2018.

\bibitem[Manakul et~al.(2023)Manakul, Liusie, and Gales]{manakul2023selfcheckgpt}
Potsawee Manakul, Adian Liusie, and Mark Gales.
\newblock Selfcheckgpt: Zero-resource black-box hallucination detection for generative large language models.
\newblock In \emph{The 2023 Conference on Empirical Methods in Natural Language Processing}, 2023.

\bibitem[Min et~al.(2023)Min, Krishna, Lyu, Lewis, Yih, Koh, Iyyer, Zettlemoyer, and Hajishirzi]{min2023factscore}
Sewon Min, Kalpesh Krishna, Xinxi Lyu, Mike Lewis, Wen-tau Yih, Pang Koh, Mohit Iyyer, Luke Zettlemoyer, and Hannaneh Hajishirzi.
\newblock Factscore: Fine-grained atomic evaluation of factual precision in long form text generation.
\newblock In \emph{Proceedings of the 2023 Conference on Empirical Methods in Natural Language Processing}, pp.\  12076--12100, 2023.

\bibitem[Touvron et~al.(2023)Touvron, Martin, Stone, Albert, Almahairi, Babaei, Bashlykov, Batra, Bhargava, Bhosale, et~al.]{touvron2023llama}
Hugo Touvron, Louis Martin, Kevin Stone, Peter Albert, Amjad Almahairi, Yasmine Babaei, Nikolay Bashlykov, Soumya Batra, Prajjwal Bhargava, Shruti Bhosale, et~al.
\newblock Llama 2: Open foundation and fine-tuned chat models.
\newblock \emph{arXiv preprint arXiv:2307.09288}, 2023.

\bibitem[Wang et~al.(2023)Wang, Yu, Zeng, Yang, Yao, Wang, Chen, Jiang, Xie, Wang, et~al.]{wang2023pandalm}
Yidong Wang, Zhuohao Yu, Zhengran Zeng, Linyi Yang, Wenjin Yao, Cunxiang Wang, Hao Chen, Chaoya Jiang, Rui Xie, Jindong Wang, et~al.
\newblock Pandalm: An automatic evaluation benchmark for llm instruction tuning optimization.
\newblock In \emph{The Twelfth International Conference on Learning Representations}, 2023.

\bibitem[Yang et~al.(2023)Yang, Sun, and Wan]{yang2023new}
Shiping Yang, Renliang Sun, and Xiaojun Wan.
\newblock A new benchmark and reverse validation method for passage-level hallucination detection.
\newblock In \emph{Findings of the Association for Computational Linguistics: EMNLP 2023}, pp.\  3898--3908, 2023.

\bibitem[Yin et~al.(2023)Yin, Sun, Guo, Wu, Qiu, and Huang]{yin2023large}
Zhangyue Yin, Qiushi Sun, Qipeng Guo, Jiawen Wu, Xipeng Qiu, and Xuan-Jing Huang.
\newblock Do large language models know what they don’t know?
\newblock In \emph{Findings of the Association for Computational Linguistics: ACL 2023}, pp.\  8653--8665, 2023.

\bibitem[Yue et~al.(2023)Yue, Wang, Chen, Zhang, Su, and Sun]{yue2023automatic}
Xiang Yue, Boshi Wang, Ziru Chen, Kai Zhang, Yu~Su, and Huan Sun.
\newblock Automatic evaluation of attribution by large language models, 2023.

\bibitem[Zhang et~al.(2019)Zhang, Kishore, Wu, Weinberger, and Artzi]{zhang2019bertscore}
Tianyi Zhang, Varsha Kishore, Felix Wu, Kilian~Q Weinberger, and Yoav Artzi.
\newblock Bertscore: Evaluating text generation with bert.
\newblock In \emph{International Conference on Learning Representations}, 2019.

\bibitem[Zhang et~al.(2023)Zhang, Li, Cui, Cai, Liu, Fu, Huang, Zhao, Zhang, Chen, et~al.]{zhang2023siren}
Yue Zhang, Yafu Li, Leyang Cui, Deng Cai, Lemao Liu, Tingchen Fu, Xinting Huang, Enbo Zhao, Yu~Zhang, Yulong Chen, et~al.
\newblock Siren's song in the ai ocean: a survey on hallucination in large language models.
\newblock \emph{arXiv preprint arXiv:2309.01219}, 2023.

\bibitem[Zhao et~al.(2023{\natexlab{a}})Zhao, Zhou, Li, Tang, Wang, Hou, Min, Zhang, Zhang, Dong, et~al.]{zhao2023survey}
Wayne~Xin Zhao, Kun Zhou, Junyi Li, Tianyi Tang, Xiaolei Wang, Yupeng Hou, Yingqian Min, Beichen Zhang, Junjie Zhang, Zican Dong, et~al.
\newblock A survey of large language models.
\newblock \emph{arXiv preprint arXiv:2303.18223}, 2023{\natexlab{a}}.

\bibitem[Zhao et~al.(2023{\natexlab{b}})Zhao, Yan, Sun, Xing, Meng, Wang, Cheng, Ren, and Yin]{zhao2023knowing}
Yukun Zhao, Lingyong Yan, Weiwei Sun, Guoliang Xing, Chong Meng, Shuaiqiang Wang, Zhicong Cheng, Zhaochun Ren, and Dawei Yin.
\newblock Knowing what llms do not know: A simple yet effective self-detection method.
\newblock \emph{arXiv preprint arXiv:2310.17918}, 2023{\natexlab{b}}.

\bibitem[Zheng et~al.(2024)Zheng, Chiang, Sheng, Zhuang, Wu, Zhuang, Lin, Li, Li, Xing, et~al.]{zheng2024judging}
Lianmin Zheng, Wei-Lin Chiang, Ying Sheng, Siyuan Zhuang, Zhanghao Wu, Yonghao Zhuang, Zi~Lin, Zhuohan Li, Dacheng Li, Eric Xing, et~al.
\newblock Judging llm-as-a-judge with mt-bench and chatbot arena.
\newblock \emph{Advances in Neural Information Processing Systems}, 36, 2024.

\bibitem[Zheng et~al.(2023)Zheng, Huang, and Chang]{zheng2023does}
Shen Zheng, Jie Huang, and Kevin Chen-Chuan Chang.
\newblock Why does chatgpt fall short in providing truthful answers.
\newblock \emph{ArXiv preprint, abs/2304.10513}, 2023.

\bibitem[Zhong et~al.(2020)Zhong, Xu, Tang, Xu, Duan, Zhou, Wang, and Yin]{zhong2020reasoning}
Wanjun Zhong, Jingjing Xu, Duyu Tang, Zenan Xu, Nan Duan, Ming Zhou, Jiahai Wang, and Jian Yin.
\newblock Reasoning over semantic-level graph for fact checking.
\newblock In \emph{Proceedings of the 58th Annual Meeting of the Association for Computational Linguistics}. Association for Computational Linguistics, 2020.

\end{thebibliography}

\clearpage

\appendix

\section{Implementation Details of HalluJudge}
\label{sec: appendix_hallujudge_config}

HalluJudge is built on the Llama-2-7B backbone \citep{touvron2023llama}. Training is conducted using 8 NVIDIA A800-SXM4-80GB GPUs, with a batch size of 1 per GPU and a maximum token truncation length of 1280. The AdamW optimizer \citep{loshchilov2018decoupled} with a learning rate of 3e-5 and a cosine learning rate scheduler is employed during training. The model is trained with bfloat16 (BF16) computation precision for two epochs, with a warmup ratio of 0.03. HalluDial is randomly split into training and test sets in a 6:4 ratio, and the average results are reported on the test set after completing training from three runs with different random seeds.

\section{Detailed Evaluation Results}

\subsection{Detailed Evaluation Results of Hallucination Detection Capabilities}
\label{sec: appendix_hallu_detect}

\begin{figure}[!h]
\centering
\includegraphics[width=\textwidth]{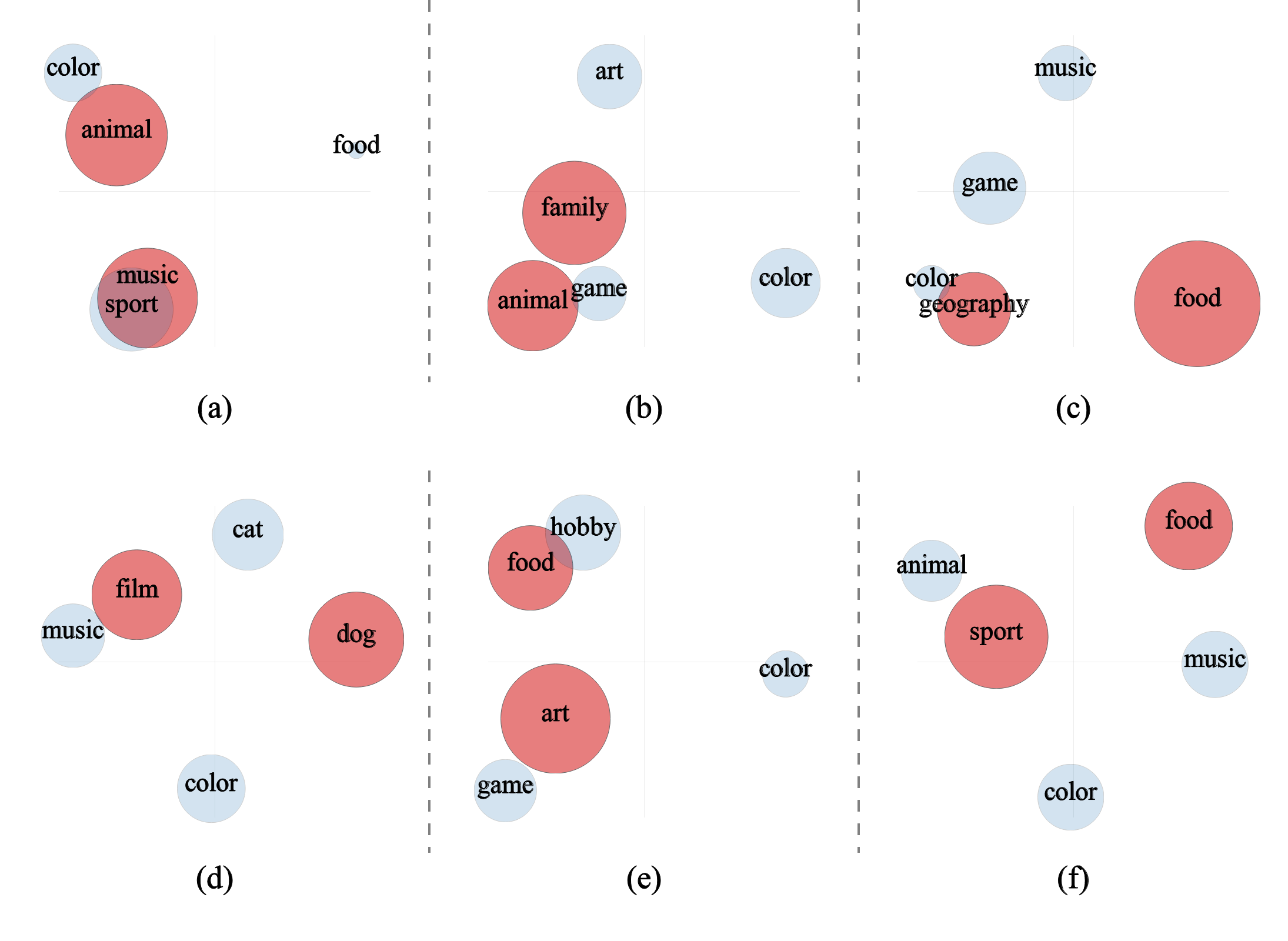}
\caption{Topic distributions of instances where LLMs fail to detect hallucinations. (a): GPT-4o-2024-05-13. (b): GPT-4-0125-preview. (c): HalluJudge. (d): GPT-3.5-turbo. (e): Llama-2-70B-chat. (f): vicuna-33B-v1.3.}
\label{fig:appendix_topic_2}
\end{figure}

Table \ref{tab:appendix_halludial_detect} presents the detailed results of various LLMs in detecting hallucinations on the HalluDial test set. The results demonstrate that HalluJudge outperforms all other LLMs across all evaluation metrics, showcasing its superior hallucination detection capabilities. Figure \ref{fig:appendix_topic_2} illustrates some of the topics in HalluDial where LLMs fail to detect hallucinations.
Furthermore, Table \ref{tab:appendix_halludial_detect_spontaneous_induced} provides a detailed breakdown of the LLMs' performance in the spontaneous and induced hallucination scenarios within the HalluDial test set. Interestingly, most LLMs exhibit better performance in the induced hallucination scenario compared to the spontaneous one. This observation suggests that the spontaneous hallucination scenario poses a more significant challenge for these models, potentially due to its diversity and complexity.
HalluJudge, in particular, excels in the induced hallucination scenario, highlighting its strong ability to identify intentionally generated hallucinated content. However, its performance in the spontaneous scenario, while still competitive, leaves room for improvement. This discrepancy could be attributed to HalluJudge's relatively small parameter scale, indicating the need for further enhancements to tackle the challenges posed by spontaneous hallucinations more effectively.
These findings underscore the complexity of the HalluDial dataset and the varying degrees of difficulty it presents for current LLMs in detecting hallucinations. The results also emphasize the importance of developing a more robust and capable judge model that can handle both spontaneous and induced hallucinations with high accuracy and reliability.

\begin{table}[!h]
\centering
\caption{Hallucination detection performance on the HalluDial test set.}
\begin{tabular}{lccccccc}
\toprule
 & Hallu. & Non-Hallu. & \multicolumn{4}{c}{Average}\\
 \cmidrule(lr){2-2} \cmidrule(lr){3-3} \cmidrule(lr){4-7}
 & F1 Score & F1 Score & Acc. & P & R & Macro F1\\
\midrule
Llama-2-7B-chat & 57.07 & 0.12 & 39.95 & 36.90 & 49.94 & 28.59\\
Llama-2-13B-chat & 57.41 & 11.64 & 42.52 & 57.52 & 51.52 & 34.53\\
Llama-2-70B-chat & 59.37 & 60.93 & 60.16 & 62.09 & 62.26 & 60.15\\
vicuna-7B-v1.5 & 19.41 & 76.44 & 63.54 & 73.00  & 54.77 & 47.92\\
vicuna-13B-v1.5 & 31.25 & 78.30 & 67.01 & 79.22 & 58.96 & 54.77\\
vicuna-33B-v1.3 & 56.26 & 61.63 & 59.12 & 59.86 & 60.23 & 58.94\\
Baichuan2-7B-Chat & 41.49 & 73.00 & 63.05 & 60.79 & 58.00 & 57.25\\
Baichuan2-13B-Chat & 62.14 & 69.59 & 66.27 & 66.09 & 66.74 & 65.87\\
internlm2-chat-7B & 43.74 & 79.85 & 70.33 & 78.93 & 63.41 & 61.80\\
internlm2-chat-20B & 50.34 & 78.63 & 70.12 & 71.96 & 64.74 & 64.49\\
chatglm3-6B & 34.31 & 72.38 & 61.11 & 57.97 & 55.15 & 53.35\\
deepseek-llm-7B-chat & 57.85 & 33.93 & 48.54 & 58.45 & 55.17 & 45.89\\
deepseek-llm-67B-chat & 50.75 & 81.92 & 73.55 & 82.95 & 66.78 & 66.33\\
Qwen1.5-7B-Chat & 54.51 & 76.25 & 68.79 & 67.75 & 65.11 & 65.38\\
Qwen1.5-32B-Chat & 50.89 & 81.51 & 73.14 & 82.02 & 66.74 & 66.20\\
Qwen1.5-72B-Chat & 49.02 & 81.38 & 72.73 & 82.98 & 66.06 & 65.20\\
GPT-3.5-turbo & 54.32 & 81.43 & 73.60 & 79.29 & 67.87 & 67.88\\
GPT-4-1106-preview & 67.75 & 85.02 & 79.54 & 83.75 & 75.24 & 76.39\\
GPT-4-0125-preview & 74.15 & 86.18 & 81.99 & 83.43 & 79.09 & 80.16\\
GPT-4o-2024-05-13 & 72.66 & 85.01 & 80.64 & 81.42 & 77.92 & 78.84\\
HalluJudge & \textbf{80.07} & \textbf{89.77} & \textbf{86.48} & \textbf{89.89} & \textbf{83.38} & \textbf{84.92}\\
\bottomrule
\end{tabular}
\label{tab:appendix_halludial_detect}
\end{table}

\begin{table}[!h]
\caption{Fine-grained hallucination detection performance in the spontaneous and induced hallucination scenarios within the HalluDial test set.}
\centering
\begin{adjustbox}{width=\textwidth}
\begin{tabular}{lcccccccc}
\toprule
 & \multicolumn{4}{c}{Spontaneous Hallucination Scenario} & \multicolumn{4}{c}{Induced Hallucination Scenario}\\
\cmidrule(lr){2-5} \cmidrule(lr){6-9}
 & Acc. & Avg.P & Avg.R & Macro F1 & Acc. & Avg.P & Avg.R & Macro F1\\
\midrule
Llama-2-7B-chat & 23.81 & 40.27 & 49.95 & 19.28 & 66.84 & 35.25 & 49.92 & 40.08\\
Llama-2-13B-chat & 27.66 & 49.33 & 49.74 & 25.76 & 67.11 & 63.18 & 50.40 & 41.29\\
Llama-2-70B-chat & 57.95 & 54.91 & 56.66 & 53.11 & 63.84 & 55.29 & 53.62 & 52.82\\
vicuna-7B-v1.5 & 75.97 & 54.32 & 50.14 & 43.94 & 42.81 & 61.77 & 56.03 & 40.42\\
vicuna-13B-v1.5 & 76.17	& 62.03 & 50.17 & 43.72 & 51.74 & 67.53 & 63.10 & 51.02\\
vicuna-33B-v1.3 & 58.24 & 53.07 & 52.35 & 51.34 & 60.58 & 51.39 & 51.06 & 50.35\\
GPT-3.5-turbo & 77.17 & 69.21 & 54.83 & 53.46 & 67.63 & 72.15 & 73.93 & 67.49\\
GPT-4-1106-preview & 82.15 & 79.54 & 66.46 & 69.25 & 75.20 & 77.87 & 80.96 & 74.93\\
GPT-4-0125-preview & \textbf{82.39} & 77.31 & \textbf{69.81} & \textbf{72.18} & 81.30 & 80.33 & 84.12 & 80.51\\
GPT-4o-2024-05-13 & 81.17 & 74.73 & 68.76 & 70.74 & 79.75 & 78.40 & 81.79 & 78.75\\
HalluJudge & 79.42 & \textbf{80.05} & 58.26 & 58.60 & \textbf{98.26} & \textbf{97.56} & \textbf{98.60} & \textbf{98.05}\\
\bottomrule
\end{tabular}
\end{adjustbox}
\label{tab:appendix_halludial_detect_spontaneous_induced}
\end{table}

\subsection{Detailed Evaluation Results of LLM's Hallucinations}
\label{sec: appendix_hallu_rate}

\begin{figure}[!h]
\centering
\includegraphics[width=0.9\textwidth]{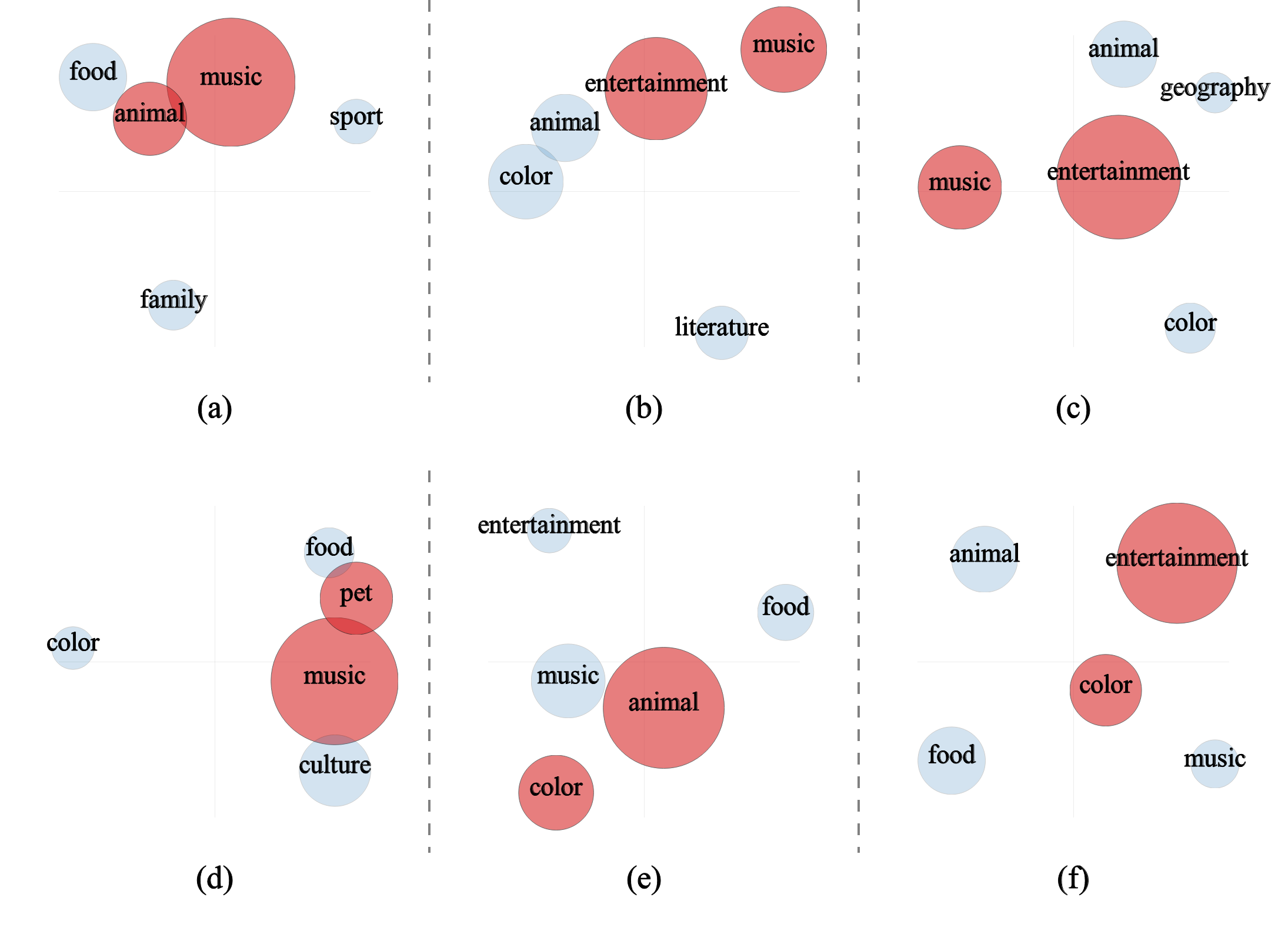}
\caption{Topic distributions of instances where LLMs are prone to hallucinations. (a): GPT-4o-2024-05-13. (b): GPT-4-0125-preview. (c): GPT-4-1106-preview. (d): GPT-3.5-turbo. (e): Llama-2-70B-chat. (f): vicuna-33B-v1.3.}
\label{fig:appendix_topic_3}
\end{figure}

Table \ref{tab: appendix_hallu_rate} presents the full evaluation results of LLM's hallucinations. Figure \ref{fig:appendix_topic_3} illustrates some of the topics in HalluDial where LLMs are prone to hallucinations.

\begin{table}[!h]
\caption{Hallucination rates of LLMs.}
\centering
\begin{tabular}{lcccc}
\toprule
 & \multicolumn{2}{c}{Gold.} & \multicolumn{2}{c}{Self.}\\
\cmidrule(lr){2-3} \cmidrule(lr){4-5}
 & turn. & dialogue. & turn. & dialogue.\\
\midrule
Llama-2-7B-chat	& 10.87 & 36.96 & 10.62 & 36.68\\
Llama-2-13B-chat & 10.10 & 35.11 & 9.09 & 32.39\\
Llama-2-70B-chat & 9.81 & 34.35 & 8.52 & 31.14\\
vicuna-7B-v1.5 & 4.39 & 17.68 & 4.31 & 17.39\\
vicuna-13B-v1.5 & 7.01 & 26.32 & 6.40 & 24.81\\
vicuna-33B-v1.3 & 6.10 & 23.97 & 5.08 & 20.14\\
Baichuan2-7B-Chat & 5.31 & 20.75 & 5.56 & 21.80\\
Baichuan2-13B-Chat & 13.20 & 45.95 & 12.37 & 43.87\\
internlm2-chat-7B & 5.82 & 22.33 & 5.08 & 20.11\\
internlm2-chat-20B & 5.41 & 20.80 & 4.61 & 18.59\\
chatglm3-6B & 5.67 & 21.92 & 5.56 & 21.86\\
deepseek-llm-7B-chat & 4.88 & 19.38 & 4.54 & 18.08\\
deepseek-llm-67B-chat & 3.91 & 15.86 & 3.30 & 13.63\\
Qwen1.5-7B-Chat & 13.38 & 45.67 & 11.43 & 40.98\\
Qwen1.5-32B-Chat & 8.87 & 32.80 & 8.70 & 32.38\\
Qwen1.5-72B-Chat & 9.50 & 34.32 & 7.67 & 28.84\\
Mistral-7B-Instruct-v0.2 & 3.14 & 12.91 & 2.74 & 11.51\\
GPT-3.5-turbo & 3.79 & 13.43 & 3.17 & 12.62\\
GPT-4-1106-preview & 5.52 & 22.72 & 5.52 & 22.33\\
GPT-4-0125-preview & 8.85 & 32.90 & 6.70 & 26.01\\
GPT-4o-2024-05-13 & 2.77 & 11.70 & 2.53 & 10.75\\
\bottomrule
\end{tabular}
\label{tab: appendix_hallu_rate}
\end{table}

\section{Can LLMs Distinguish the Hallucinatory Response in a Comparative Scenario?}

To further investigate the ability of LLMs to identify hallucinations, we test their performance in a comparative setting within the same conversational context. Specifically, we simultaneously provide the models with two responses: one containing hallucinations and another without. The data used in this experiment is derived from the induced hallucination scenario, and the LLMs are guided to perform the hallucination identification task using the instructions shown in Table \ref{tab: inst_hallu_detect_comparative}. 
Notably, considering the position bias in LLMs, we alternate the order of the two responses for each model and perform inference twice to derive the final results. In addition to accuracy, we also report the invalid rate for cases where the output fails to meet the required criteria.
As shown in Table \ref{tab: appendix_hallu_detect_comparison}, in terms of average accuracy, the results reveal a polarized trend in the performance of LLMs. Models with larger parameter scales demonstrate a remarkable ability to accurately identify the hallucinatory response from the two provided options. Conversely, smaller models exhibit poor performance under this setting, with most of their results closely approximating random selection. 
Furthermore, we observe a severe position bias in most of the open-source LLMs, where they tend to favor either the former or latter response regardless of its content. For instance, vicuna-7B-v1.5 and Baichuan2-13B-Chat show a strong preference for the first response, while vicuna-33B-v1.3 and deepseek-llm-7B-chat heavily favor the second one. This finding suggests that these models may lack a genuine understanding of hallucinations and instead rely on superficial cues to make their decisions.

\begin{table}[!h]
\caption{Results of hallucination detection in the comparative setting. Accuracy\textsubscript{F} denotes the former response is hallucinatory, while Accuracy\textsubscript{L} represents the latter one contains hallucinations.}
\centering
\begin{tabular}{lcccc}
\toprule
 & Accuracy\textsubscript{F} &  Accuracy\textsubscript{L} & Accuracy\textsubscript{average} & Invalid Rate\\
\midrule
vicuna-7B-v1.5 & 97.63 & 4.64 & 51.13 & -\\
vicuna-13B-v1.5 & 70.12 & 23.83 & 46.97 & -\\
vicuna-33B-v1.3 & 7.58 & 98.05 & 53.29 & -\\
Mistral-7B-Instruct-v0.2 & 65.06 & 31.68 & 48.32 & 28.64\\
Baichuan2-7B-Chat & 49.09 & 17.76 & 32.93 & 20.89\\
Baichuan2-13B-Chat & 99.56 & 3.12 & 51.34 & -\\
internlm2-chat-7B & 80.70 & 38.04 & 59.37 & -\\
internlm2-chat-20B & 51.81 & 82.87 & 67.34 & -\\
deepseek-llm-7B-chat & 0.28 & 99.83 & 50.05 & -\\
deepseek-llm-67B-chat & 70.27 & 92.38 & 81.33 & -\\
Qwen1.5-7B-Chat & 74.85 & 44.91 & 59.88 & -\\
Qwen1.5-32B-Chat & 80.40 & 52.54 & 66.47 & -\\
Qwen1.5-72B-Chat & 95.35 & 65.36 & 80.76 & 9.86\\
GPT-3.5-turbo-1106 & 70.33 & 73.40 & 71.87 & -\\
GPT-4-1106-preview & 90.56 & 85.89 & 88.20 & 3.26\\

\bottomrule
\end{tabular}
\label{tab: appendix_hallu_detect_comparison}
\end{table}

\section{Annotation Details}
\label{appendix: annotation_details}

To evaluate the performance of GPT-3.5-turbo, GPT-4-1106-preview, and HalluJudge, we initially sample 850 instances where the models' hallucination detection decisions are hallucinatory. Subsequently, we instruct three annotators to assess the reasonableness and accuracy of the generated localization and explanation. The annotators consist of two human labelers and GPT-4o-2024-05-13. The instruction for GPT-4o-2024-05-13 is shown in Table \ref{tab: inst_hallu_evaluate_annotate}. To measure inter-annotator agreement (IAA), we employ Cohen's Kappa ($\kappa$) and obtain an average $\kappa$ value of 0.902 ($0.80 \leq \kappa \leq 1.00$), indicating perfect consistency among the annotators.

\section{More Details on HalluDial}

\begin{figure}[!h]
\centering
\includegraphics[width=\textwidth]{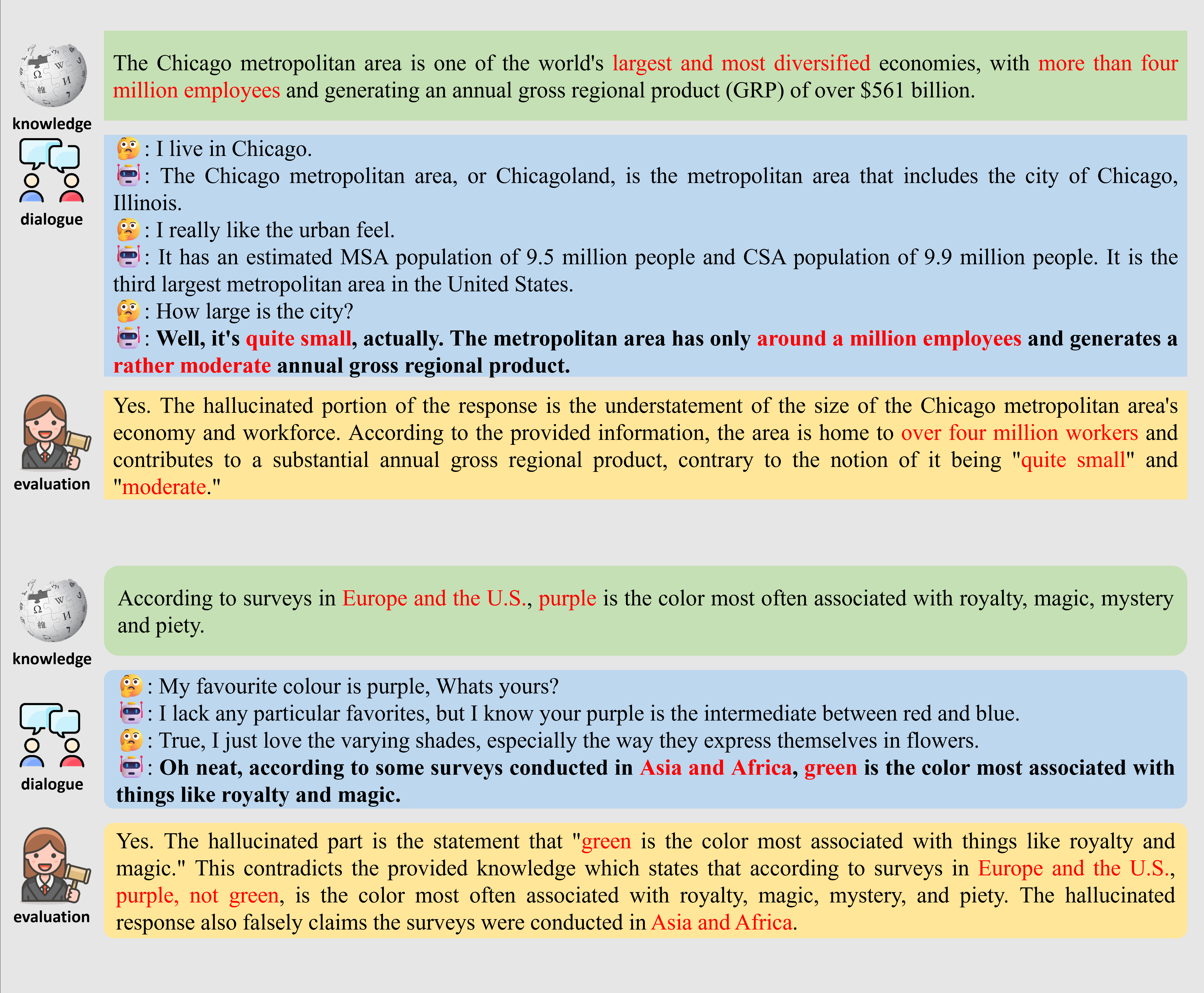}
\caption{Factuality hallucination examples of HalluDial.}
\label{fig:appendix_detailsamples_factuality}
\end{figure}

\begin{figure}[!h]
\centering
\includegraphics[width=\textwidth]{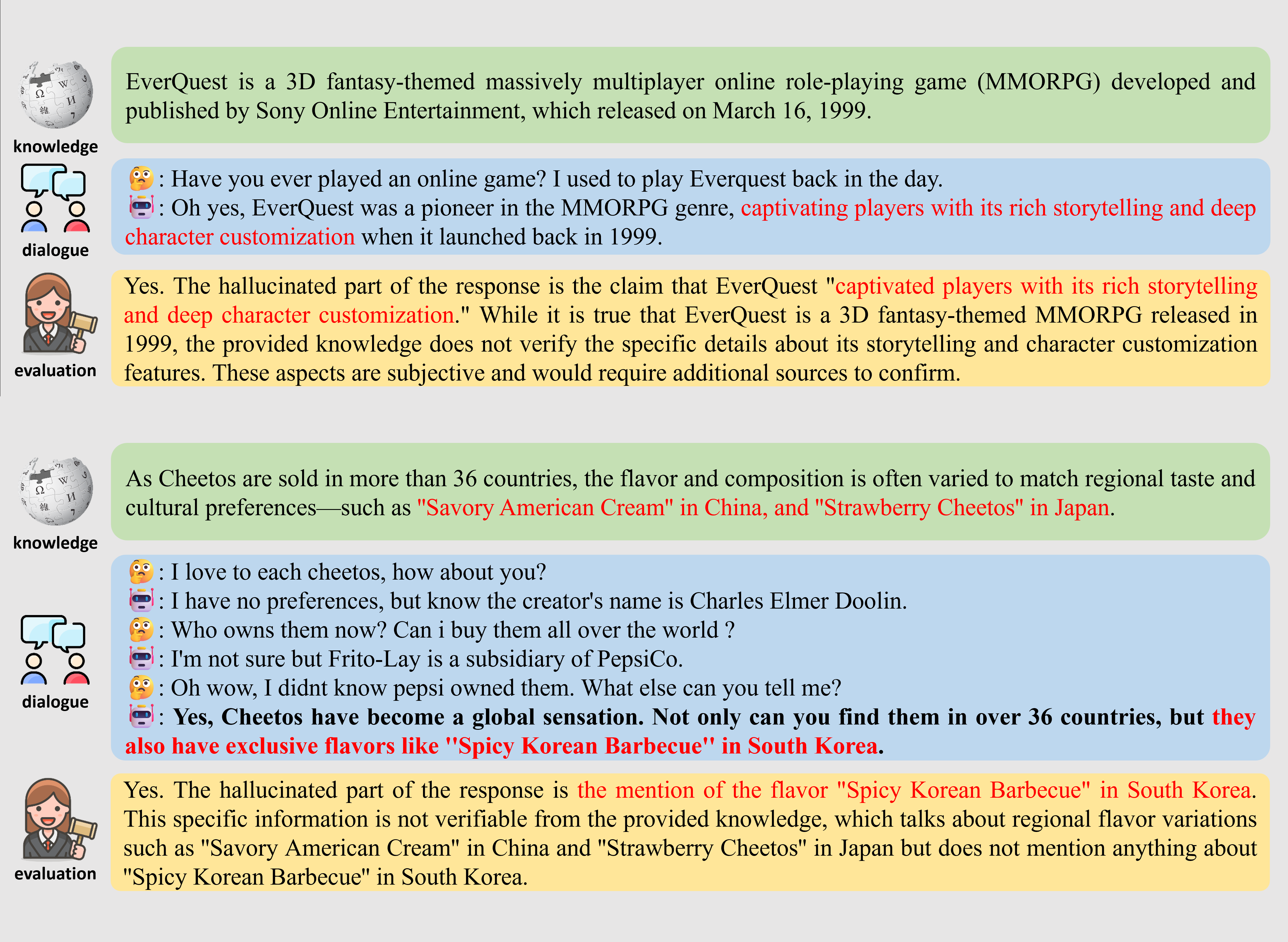}
\caption{Faithfulness hallucination examples of HalluDial.}
\label{fig:appendix_detailsamples_faithfulness}
\end{figure}

\begin{figure}[!h]
\centering
\includegraphics[width=\textwidth]{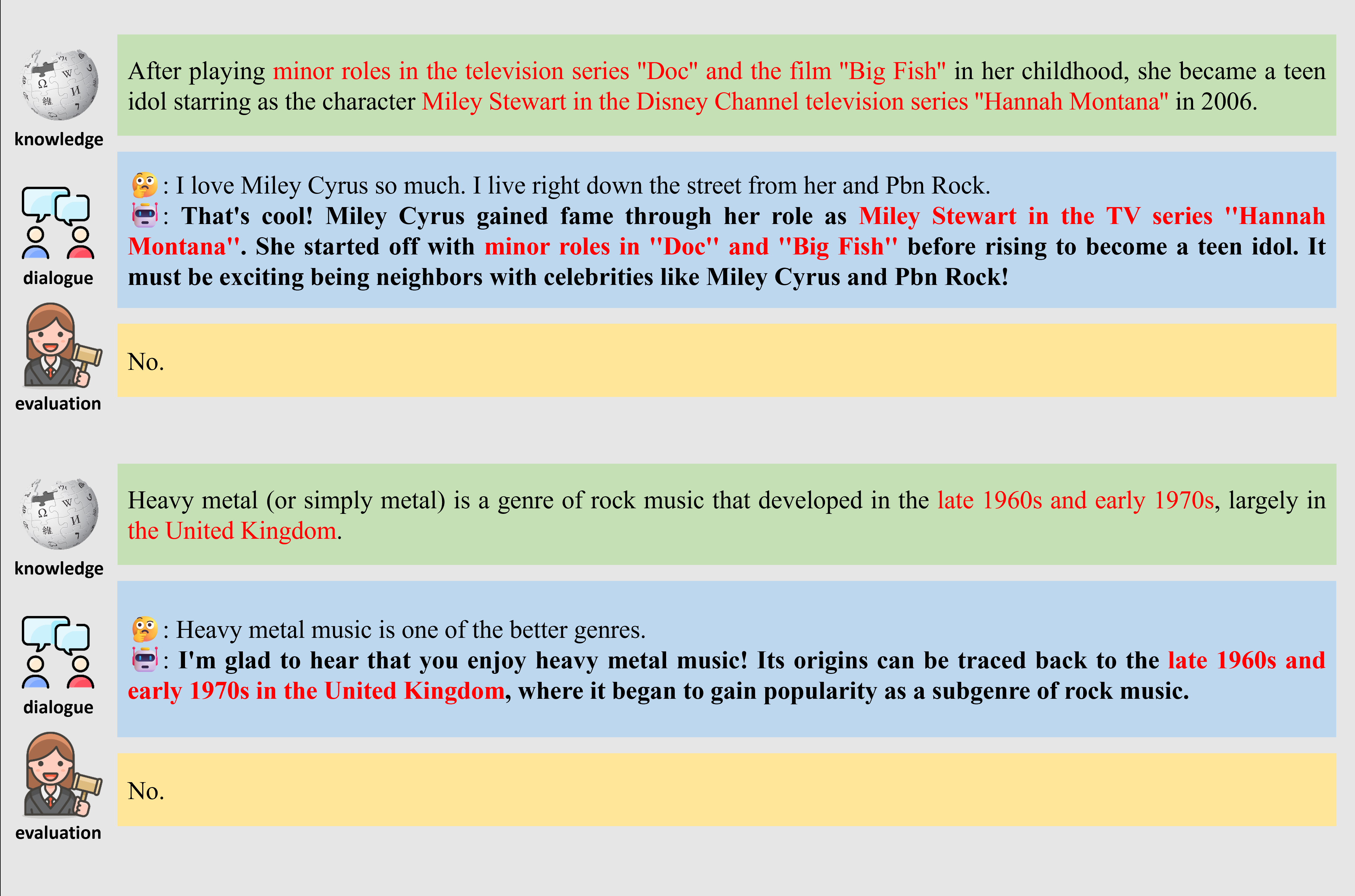}
\caption{Non-hallucinatory examples of HalluDial.}
\label{fig:appendix_detailsamples_non-hallucination}
\end{figure}

We provide more examples of HalluDial in Figure \ref{fig:appendix_detailsamples_factuality}, Figure \ref{fig:appendix_detailsamples_faithfulness}, and Figure \ref{fig:appendix_detailsamples_non-hallucination}.

\clearpage

\section{Prompt Template}

\subsection{Instructions in the Spontaneous Scenario}

\begin{table}[!htb]
\caption{Instruction of diverse dialogue sampling.}
\centering
\begin{tabular}{p{\textwidth}}
\toprule
\rowcolor{Mycolor1!60} As an assistant, your task is to engage in a knowledge-grounded conversation. You are required to compose a SINGLE response in English, not exceeding 80 words, based on the provided knowledge and consistent with the previous dialogue history. Do NOT add extra preamble or explanations and refrain from generating follow-up prompts or multiple dialogue turns. Ensure your response is coherent and do not make claims that cannot be supported by the given knowledge and conversation context.
\\
\\

\textbf{\#\#\#Knowledge\#\#\#}: <Here is the provided external knowledge>
\\
\textbf{\#\#\#Dialogue History\#\#\#}: <Here is the dialogue history>
\\
\bottomrule
\end{tabular}
\label{tab: diverese_dialogue_sampling}
\end{table}

\begin{table}[!htb]
\caption{Instruction of automatic hallucination annotation.}
\begin{tabular}{p{\textwidth}}
\toprule
\rowcolor{Mycolor1!60} As an assistant, your task is to serve as a response judge. Provided with the related knowledge, a dialogue history, a reference response and a generated response, your objective is to determine if the generated response contains any hallucinated information that (1) directly contradicts the given knowledge, or (2) cannot be verified from the provided knowledge and dialogue context.
\\
\\
\rowcolor{Mycolor2!80}You SHOULD refer to the following examples:\\
\rowcolor{Mycolor2!80}\\
\rowcolor{Mycolor2!80} Example 1:
\\
\rowcolor{Mycolor2!80} \textbf{\#\#\#Knowledge\#\#\#}: Use by a wider audience only came in 1995 when restrictions on the use of the Internet to carry commercial traffic were lifted.\\
\rowcolor{Mycolor2!80} \textbf{\#\#\#Dialogue History\#\#\#}: [Human]: Can you imagine the world without internet access? [Assistant]: Yeah, but once the access to the internet was a rare thing. do you remember? [Human]: I do. What else can you tell me ?\\
\rowcolor{Mycolor2!80} \textbf{\#\#\#Reference Response\#\#\#}: Well, I know that more people started using it after some restrictions on internet use were lifted in 1995.\\
\rowcolor{Mycolor2!80} \textbf{\#\#\#Generated Response\#\#\#}: Well, I know that the internet was primarily available to the public in the early 1980s, with no restrictions on commercial traffic from the very beginning. \\
\rowcolor{Mycolor2!80} \textbf{\#\#\#Judgement\#\#\#}: Yes. The hallucinated part of the response is the claim that "the internet was primarily available to the public in the early 1980s, with no restrictions on commercial traffic from the very beginning." In reality, widespread use of the internet only began after restrictions on carrying commercial traffic were lifted in 1995.\\
\rowcolor{Mycolor2!80} \\
\rowcolor{Mycolor2!80} <More In-Context Examples>
\\
\\
\rowcolor{Mycolor3!60} You SHOULD first carefully evaluate each segment of the generated response and determine if the response contains any hallucinated information. If any such content is found, your next step is to identify the specific hallucinated segment and explain why it qualifies as such. Your judgement MUST strictly follow the format demonstrated in the examples above and be well-organized, clear and concise.
\\
\\
\textbf{\#\#\#Knowledge\#\#\#}: <Here is the provided external knowledge>\\
\textbf{\#\#\#Dialogue History\#\#\#}: <Here is the dialogue history>\\
\textbf{\#\#\#Reference Response\#\#\#}: <Here is the reference response>\\
\textbf{\#\#\#Generated Response\#\#\#}: <Here is the generated response>\\
\textbf{\#\#\#Judgement\#\#\#}: \\
\bottomrule
\end{tabular}
\label{tab: automatic_hallu_annotation}
\end{table}

\clearpage

\subsection{Instructions in the Induced Scenario}

\begin{table}[!htb]
\caption{Instruction of generating factuality hallucination samples.}
\begin{tabular}{p{\textwidth}}
\toprule
\rowcolor{Mycolor1!60} As an assistant, your task is to generate a hallucinated response within the context of a conversation that contradicts the related knowledge. Provided with the related knowledge, a dialogue history, and a reference response grounded in the knowledge, your objectives are to:
(1) Generate a hallucinated response that contradicts the related knowledge. 
(2) Identify the specific part of the response that is hallucinated and explain the reasons.
\\
\\
\rowcolor{Mycolor2!80}You SHOULD refer to the following examples:\\
\rowcolor{Mycolor2!80}\\
\rowcolor{Mycolor2!80} Example 1:\\
\rowcolor{Mycolor2!80} \textbf{\#\#\#Knowledge\#\#\#}: Use by a wider audience only came in 1995 when restrictions on the use of the Internet to carry commercial traffic were lifted.\\
\rowcolor{Mycolor2!80} \textbf{\#\#\#Dialogue History\#\#\#}: [Human]: Can you imagine the world without internet access? [Assistant]: Yeah, but once the access to the internet was a rare thing. do you remember? [Human]: I do. What else can you tell me ?\\
\rowcolor{Mycolor2!80} \textbf{\#\#\#Reference Response\#\#\#}: Well, I know that more people started using it after some restrictions on internet use were lifted in 1995.\\
\rowcolor{Mycolor2!80} \textbf{\#\#\#Hallucinated Response\#\#\#}: Well, I know that the internet was primarily available to the public in the early 1980s, with no restrictions on commercial traffic from the very beginning.  \\
\rowcolor{Mycolor2!80} \textbf{\#\#\#Hallucination Explanation\#\#\#}: The hallucinated part of the response is the claim that "the internet was primarily available to the public in the early 1980s, with no restrictions on commercial traffic from the very beginning." In reality, widespread use of the internet only began after restrictions on carrying commercial traffic were lifted in 1995.\\
\rowcolor{Mycolor2!80} \\
\rowcolor{Mycolor2!80} <More In-Context Examples>
\\
\\
\rowcolor{Mycolor3!60} The hallucinated response SHOULD appear plausible but be factually incorrect, and SHOULD NOT exceed the length of the reference response by more than six words. You SHOULD avoid deliberately emphasizing the authenticity of the generated response and strive to seamlessly align it with the tone and style of the reference response, maintaining a natural flow in the ongoing conversation.
\\
\\
\textbf{\#\#\#Knowledge\#\#\#}: <Here is the provided external knowledge>\\
\textbf{\#\#\#Dialogue History\#\#\#}: <Here is the dialogue history>\\
\textbf{\#\#\#Reference Response\#\#\#}: <Here is the reference response>\\
\bottomrule
\end{tabular}
\label{tab: gen_fact_hallu}
\end{table}

\begin{table}[!htb]
\caption{Instruction of generating faithfulness hallucination samples.}
\begin{tabular}{p{\textwidth}}
\toprule
\rowcolor{Mycolor1!60} As an assistant, your task is to generate a hallucinated response within the context of a conversation, which contains some information that cannot be verified from the related knowledge. Provided with the related knowledge, a dialogue history, and a reference response grounded in the knowledge, your objectives are to:
(1) Generate a hallucinated response containing some information that cannot be verified from the related knowledge.
(2) Identify the specific part of the response that is hallucinated and explain the reasons.
\\
\\
\rowcolor{Mycolor2!80}You SHOULD refer to the following examples:\\
\rowcolor{Mycolor2!80}\\
\rowcolor{Mycolor2!80} Example 1:\\
\rowcolor{Mycolor2!80} \textbf{\#\#\#Knowledge\#\#\#}: Use by a wider audience only came in 1995 when restrictions on the use of the Internet to carry commercial traffic were lifted.\\
\rowcolor{Mycolor2!80} \textbf{\#\#\#Dialogue History\#\#\#}: [Human]: Can you imagine the world without internet access? [Assistant]: Yeah, but once the access to the internet was a rare thing. do you remember? [Human]: I do. What else can you tell me ?\\
\rowcolor{Mycolor2!80} \textbf{\#\#\#Reference Response\#\#\#}: Well, I know that more people started using it after some restrictions on internet use were lifted in 1995.\\
\rowcolor{Mycolor2!80} \textbf{\#\#\#Hallucinated Response\#\#\#}: Yes, the internet truly came into its own when in 1995, a breakthrough in the optical fiber technology significantly increased its speed and the existing commercial traffic restrictions were also lifted.  \\
\rowcolor{Mycolor2!80} \textbf{\#\#\#Hallucination Explanation\#\#\#}: The hallucinated part of the response is the claim that "a breakthrough in the optical fiber technology significantly increased its speed." This piece of information is not verifiable from the given knowledge, which only mentions the lifting of restrictions on carrying commercial traffic in 1995. There's no mention of any technological breakthrough in optical fiber technology that improved internet speed at that time.\\
\rowcolor{Mycolor2!80} \\
\rowcolor{Mycolor2!80} <More In-Context Examples>
\\
\\
\rowcolor{Mycolor3!60} The hallucinated response SHOULD appear plausible and SHOULD NOT exceed the length of the reference response by more than six words. You SHOULD avoid deliberately emphasizing the authenticity of the generated response and strive to seamlessly align it with the tone and style of the reference response, maintaining a natural flow in the ongoing conversation.
\\
\\
\textbf{\#\#\#Knowledge\#\#\#}: <Here is the provided external knowledge>\\
\textbf{\#\#\#Dialogue History\#\#\#}: <Here is the dialogue history>\\
\textbf{\#\#\#Reference Response\#\#\#}: <Here is the reference response>\\
\bottomrule
\end{tabular}
\label{tab: gen_faith_hallu}
\end{table}

\clearpage

\subsection{Instructions Used during Evaluations}

\begin{table}[!htb]
\caption{Instruction of hallucination detection.}
\begin{tabular}{p{\textwidth}}
\toprule
\rowcolor{Mycolor1!60} As an assistant, your task is to serve as a response judge. Provided with the related knowledge, a dialogue history and a generated response, your objective is to determine if the generated response contains any hallucinated information that (1) directly contradicts the given knowledge, or (2) cannot be verified from the provided knowledge and dialogue context.
\\
\\
\rowcolor{Mycolor3!60} You SHOULD carefully evaluate each segment of the generated response and determine if the response contains any hallucinated information. If such content is detected, your output should be "Yes"; otherwise, it should be "No".
\\
\\
\textbf{\#\#\#Knowledge\#\#\#}: <Here is the provided external knowledge>\\
\textbf{\#\#\#Dialogue History\#\#\#}: <Here is the dialogue history>\\
\textbf{\#\#\#Response\#\#\#}: <Here is the response>\\
\textbf{\#\#\#Judgement\#\#\#}:\\
\bottomrule
\end{tabular}
\label{tab: inst_hallu_detect}
\end{table}

\begin{table}[!htb]
\caption{Instruction of hallucination localization and rationale provision.}
\begin{tabular}{p{\textwidth}}
\toprule
\rowcolor{Mycolor1!60} As an assistant, your task is to serve as a response judge. Provided with the related knowledge, a dialogue history and a generated response, your objective is to determine if the generated response contains any hallucinated information that (1) directly contradicts the given knowledge, or (2) cannot be verified from the provided knowledge and dialogue context.
\\
\\
\rowcolor{Mycolor2!80}You SHOULD refer to the following examples:\\
\rowcolor{Mycolor2!80}\\
\rowcolor{Mycolor2!80} Example 1:\\
\rowcolor{Mycolor2!80} \textbf{\#\#\#Knowledge\#\#\#}: Use by a wider audience only came in 1995 when restrictions on the use of the Internet to carry commercial traffic were lifted.\\
\rowcolor{Mycolor2!80} \textbf{\#\#\#Dialogue History\#\#\#}: [Human]: Can you imagine the world without internet access? [Assistant]: Yeah, but once the access to the internet was a rare thing. do you remember? [Human]: I do. What else can you tell me ?\\
\rowcolor{Mycolor2!80} \textbf{\#\#\#Response\#\#\#}: Well, I know that the internet was primarily available to the public in the early 1980s, with no restrictions on commercial traffic from the very beginning. \\
\rowcolor{Mycolor2!80} \textbf{\#\#\#Judgement\#\#\#}: Yes. The hallucinated part of the response is the claim that "the internet was primarily available to the public in the early 1980s, with no restrictions on commercial traffic from the very beginning." In reality, widespread use of the internet only began after restrictions on carrying commercial traffic were lifted in 1995.  \\
\rowcolor{Mycolor2!80} \\
\rowcolor{Mycolor2!80} <More In-Context Examples>
\\
\\
\rowcolor{Mycolor3!60} You SHOULD first carefully evaluate each segment of the generated response and determine if the response contains any hallucinated information. If any such content is found, your next step is to identify the specific hallucinated segment and explain why it qualifies as such. Your judgement MUST strictly follow the format demonstrated in the examples above and be well-organized, clear and concise.
\\
\\
\textbf{\#\#\#Knowledge\#\#\#}: <Here is the provided external knowledge>\\
\textbf{\#\#\#Dialogue History\#\#\#}: <Here is the dialogue history>\\
\textbf{\#\#\#Response\#\#\#}: <Here is the response>\\
\textbf{\#\#\#Judgement\#\#\#}:\\
\bottomrule
\end{tabular}
\label{tab: inst_hallu_rationales}
\end{table}

\begin{table}[!htb]
\caption{Instruction of hallucination detection in a comparative setting.}
\begin{tabular}{p{\textwidth}}
\toprule
\rowcolor{Mycolor1!60} As an assistant, your task is to serve as a response judge. Provided with the related knowledge, a dialogue history and two generated responses, your objective is to determine which response contains hallucinated information that (1) directly contradicts the given knowledge, or (2) cannot be verified from the provided knowledge and dialogue context.
\\
\\
\rowcolor{Mycolor3!60} You SHOULD carefully evaluate each segment of both generated responses and determine which response contains more or any hallucinated information. Your output should be strictly "1" or "2" depending on which response contains the hallucinated information. 
\\
\\
\textbf{\#\#\#Knowledge\#\#\#}: <Here is the provided external knowledge>\\
\textbf{\#\#\#Dialogue History\#\#\#}: <Here is the dialogue history>\\
\textbf{\#\#\#Response 1\#\#\#}: <Here is Response 1>\\
\textbf{\#\#\#Response 2\#\#\#}: <Here is Response 2>\\
\textbf{\#\#\#Judgement\#\#\#}:\\
\bottomrule
\end{tabular}
\label{tab: inst_hallu_detect_comparative}
\end{table}

\begin{table}[!htb]
\caption{Instruction of evaluating hallucination localization and explanation.}
\begin{tabular}{p{\textwidth}}
\toprule
\rowcolor{Mycolor1!60} As an assistant, your task is to serve as a meta-judge annotator. You will be provided with relevant knowledge, a dialogue history, a response based on this knowledge, and a judgement about whether the response contains hallucinations. Your objective is to critically evaluate the accuracy of this judgement.
\\
\\
\rowcolor{Mycolor1!60} A hallucination in this context is defined as information in the response that either (1) directly contradicts the given knowledge or (2) cannot be verified from the provided knowledge and dialogue context. If the response does not contain any hallucinated information, the accurate judgement should be "No". Conversely, if any hallucination is present, the accurate judgement should be "Yes", and it should identify the specific hallucinated segment and provide the corresponding justifications.
\\
\\
\rowcolor{Mycolor3!60} You SHOULD carefully assess the correctness of the judgement. Your answer should strictly be either "Accurate" if the judgement correctly identified the presence or absence of hallucination, or "Inaccurate" if it did not.
\\
\\
\textbf{\#\#\#Knowledge\#\#\#}: <Here is the provided external knowledge>\\
\textbf{\#\#\#Dialogue History\#\#\#}: <Here is the dialogue history>\\
\textbf{\#\#\#Response\#\#\#}: <Here is the Response>\\
\textbf{\#\#\#Judgement\#\#\#}: <Here is the judgement>\\
\textbf{\#\#\#Your Meta-Judgement\#\#\#}:\\
\bottomrule
\end{tabular}
\label{tab: inst_hallu_evaluate_annotate}
\end{table}

\clearpage

\section{Limitations and Negative Societal Impacts}
\label{sec: appendix_limitations}

We acknowledge that there are some limitations in our research. Firstly, since some of the responses in our dialogues are generated by GPT-4 and other large language models, and labeled by GPT-4, the dataset may inherit biases from these models to a certain extent. Although we cannot completely eliminate these biases, we have taken measures to mitigate their impact, such as utilizing multiple publicly available large language models to generate responses, thereby enhancing the diversity and representativeness of the data. Secondly, while the HalluDial dataset is relatively large and covers multiple domains, it may not fully capture all real-world conversation scenarios. 
Our dataset has minimal potential for negative societal impact. This is primarily because our data is sourced from publicly available seed datasets, and widely-used large language models (e.g., GPT-4), which inherently limits the risk of adverse effects on individuals or society as a whole.

\section{Responsibility and Dataset Liscence}
\label{sec: appendix_liscence}

We bear all responsibility in case of violation of rights and our dataset is under the license of CC BY-NC-SA.

\section{Datasheets for Our Dataset}
\label{sec: appendix_datasheet}

\subsection{Motivation}

\begin{itemize}

\item \textbf{For what purpose was the dataset created?} Was there a specific task in mind? Was there a specific gap that needed to be filled? Please provide a description.

This dataset was created to study and evaluate hallucinations of large language models in information-seeking dialogues, a specific gap is mentioned in the Introduction.

\item \textbf{Who created the dataset (e.g., which team, research group) and on behalf of which entity (e.g., company, institution, organization)?}

This dataset was created by the authors of this paper.

\item \textbf{Who funded the creation of the dataset?} If there is an associated grant, please provide the name of the grantor and the grant name and number.

The institute of the authors funded the creation of the dataset.

\item \textbf{Any other comments?}

None.

\end{itemize}

\subsection{Composition}

\begin{itemize}

\item \textbf{What do the instances that comprise the dataset
    represent (e.g., documents, photos, people, countries)?} Are there
  multiple types of instances (e.g., movies, users, and ratings;
  people and interactions between them; nodes and edges)? Please
  provide a description.

An instance of our dataset represents a dialogue. The description is provided in our paper.

\item \textbf{How many instances are there in total (of each type, if appropriate)?}

Our dataset includes 4094 dialogues.

\item \textbf{Does the dataset contain all possible instances or is it
    a sample (not necessarily random) of instances from a larger set?}
  If the dataset is a sample, then what is the larger set? Is the
  sample representative of the larger set (e.g., geographic coverage)?
  If so, please describe how this representativeness was
  validated/verified. If it is not representative of the larger set,
  please describe why not (e.g., to cover a more diverse range of
  instances, because instances were withheld or unavailable).

No.

\item \textbf{What data does each instance consist of?} 

It is mentioned in this paper.

\item \textbf{Is there a label or target associated with each
    instance?} If so, please provide a description.

Yes. The description is provided in this paper.

\item \textbf{Is any information missing from individual instances?}
  If so, please provide a description, explaining why this information
  is missing (e.g., because it was unavailable). This does not include
  intentionally removed information, but might include, e.g., redacted
  text.

No.

\item \textbf{Are relationships between individual instances made
    explicit (e.g., users' movie ratings, social network links)?} If
  so, please describe how these relationships are made explicit.

No. Instances concentrate on the same phenomenon.

\item \textbf{Are there recommended data splits (e.g., training,
    development/validation, testing)?} If so, please provide a
  description of these splits, explaining the rationale behind them.

Yes. It is provided in Appendix \ref{sec: appendix_hallujudge_config}.

\item \textbf{Are there any errors, sources of noise, or redundancies
    in the dataset?} If so, please provide a description.

No, the dataset is automatically generated by widely-used open-source large language models or API-based large language models (e.g., GPT-4). 

\item \textbf{Is the dataset self-contained, or does it link to or
    otherwise rely on external resources (e.g., websites, tweets,
    other datasets)?} If it links to or relies on external resources,
    a) are there guarantees that they will exist, and remain constant,
    over time; b) are there official archival versions of the complete
    dataset (i.e., including the external resources as they existed at
    the time the dataset was created); c) are there any restrictions
    (e.g., licenses, fees) associated with any of the external
    resources that might apply to a dataset consumer? Please provide
    descriptions of all external resources and any restrictions
    associated with them, as well as links or other access points, as
    appropriate.

It’s self-contained.

\item \textbf{Does the dataset contain data that might be considered
    confidential (e.g., data that is protected by legal privilege or
    by doctor-patient confidentiality, data that includes the content of individuals' non-public communications)?} If so, please provide
    a description.

No.

\item \textbf{Does the dataset contain data that, if viewed directly,
    might be offensive, insulting, threatening, or might otherwise
    cause anxiety?} If so, please describe why.

Yes. Some of the dialogues are about big events. Thus they may be offensive for people. However, we believe our dataset's offensiveness to be minimal, as the sources of the dataset are publicly available seed datasets and widely-used large language models.

\end{itemize}

\begin{itemize}

\item \textbf{Does the dataset identify any subpopulations (e.g., by
    age, gender)?} If so, please describe how these subpopulations are
  identified and provide a description of their respective
  distributions within the dataset.

No.

\item \textbf{Is it possible to identify individuals (i.e., one or
    more natural persons), either directly or indirectly (i.e., in
    combination with other data) from the dataset?} If so, please
    describe how.

Yes. Their names are given in the dialogues.

\item \textbf{Does the dataset contain data that might be considered
    sensitive in any way (e.g., data that reveals race or ethnic
    origins, sexual orientations, religious beliefs, political
    opinions or union memberships, or locations; financial or health
    data; biometric or genetic data; forms of government
    identification, such as social security numbers; criminal
    history)?} If so, please provide a description.

Yes. Our dataset may contain dialogues about religious beliefs, political opinions and so on.

\item \textbf{Any other comments?}

None.

\end{itemize}

\subsection{Collection Process}

\begin{itemize}

\item \textbf{How was the data associated with each instance
    acquired?} Was the data directly observable (e.g., raw text, movie
  ratings), reported by subjects (e.g., survey responses), or
  indirectly inferred/derived from other data (e.g., part-of-speech
  tags, model-based guesses for age or language)? If the data was reported
  by subjects or indirectly inferred/derived from other data, was the
  data validated/verified? If so, please describe how.

The original dialogues are from the open-source dialogue dataset. We present details in our paper.

\item \textbf{What mechanisms or procedures were used to collect the
    data (e.g., hardware apparatuses or sensors, manual human
    curation, software programs, software APIs)?} How were these
    mechanisms or procedures validated?

The software program and software APIs. We present the details in our paper.

\item \textbf{Who was involved in the data collection process (e.g.,
    students, crowdworkers, contractors) and how were they compensated
    (e.g., how much were crowdworkers paid)?}

Crowdworkers. They are paid nicely.

\item \textbf{Over what timeframe was the data collected?} Does this
  timeframe match the creation timeframe of the data associated with
  the instances (e.g., recent crawl of old news articles)?  If not,
  please describe the timeframe in which the data associated with the
  instances was created.

The dataset was collected in the early Spring of 2024. Yes.

\item \textbf{Were any ethical review processes conducted (e.g., by an
    institutional review board)?} If so, please provide a description
  of these review processes, including the outcomes, as well as a link
  or other access point to any supporting documentation.

No.

\end{itemize}

\begin{itemize}

\item \textbf{Did you collect the data from the individuals in
    question directly, or obtain it via third parties or other sources
    (e.g., websites)?}

By curating and augmenting a published dataset.

\item \textbf{Were the individuals in question notified about the data
    collection?} If so, please describe (or show with screenshots or
  other information) how notice was provided, and provide a link or
  other access point to, or otherwise reproduce, the exact language of
  the notification itself.

N/A.

\item \textbf{Did the individuals in question consent to the
    collection and use of their data?} If so, please describe (or show
  with screenshots or other information) how consent was requested and
  provided, and provide a link or other access point to, or otherwise
  reproduce, the exact language to which the individuals consented.

N/A.

\item \textbf{If consent was obtained, were the consenting individuals
    provided with a mechanism to revoke their consent in the future or
    for certain uses?} If so, please provide a description, as well as
  a link or other access point to the mechanism (if appropriate).

N/A.

\item \textbf{Has an analysis of the potential impact of the dataset
    and its use on data subjects (e.g., a data protection impact
    analysis) been conducted?} If so, please provide a description of
  this analysis, including the outcomes, as well as a link or other
  access point to any supporting documentation.

No. We believe our dataset to have a limited negative effect, for all of the data is from either published datasets or widely-used large language models.

\item \textbf{Any other comments?}

None.

\end{itemize}

\subsection{Preprocessing/cleaning/labeling}

\begin{itemize}

\item \textbf{Was any preprocessing/cleaning/labeling of the data done
    (e.g., discretization or bucketing, tokenization, part-of-speech
    tagging, SIFT feature extraction, removal of instances, processing
    of missing values)?} If so, please provide a description. If not,
  you may skip the remaining questions in this section.

No.

\end{itemize}

\subsection{Uses}

\begin{itemize}

\item \textbf{Has the dataset been used for any tasks already?} If so, please provide a description.

No.

\item \textbf{What (other) tasks could the dataset be used for?}

We provide the details in our paper.

\item \textbf{Is there anything about the composition of the dataset or the way it was collected and preprocessed/cleaned/labeled that might impact future uses?} For example, is there anything that a dataset consumer might need to know to avoid uses that could result in unfair treatment of individuals or groups (e.g., stereotyping, quality of service issues) or other risks or harms (e.g., legal risks, financial harms)? If so, please provide a description. Is there anything a dataset consumer could do to mitigate these risks or harms?

No.

\item \textbf{Are there tasks for which the dataset should not be used?} If so, please provide a description.

No.

\item \textbf{Any other comments?}

None.

\end{itemize}

\subsection{Distribution}

\begin{itemize}

\item \textbf{Will the dataset be distributed to third parties outside of the entity (e.g., company, institution, organization) on behalf of which the dataset was created?} If so, please provide a description.

Yes, the dataset is publicly available.

\item \textbf{How will the dataset will be distributed (e.g., tarball on website, API, GitHub)?} Does the dataset have a digital object identifier (DOI)?

On Github.

\item \textbf{When will the dataset be distributed?}

It’s already been distributed.

\item \textbf{Will the dataset be distributed under a copyright or other intellectual property (IP) license, and/or under applicable terms of use (ToU)?} If so, please describe this license and/or ToU, and provide a link or other access point to, or otherwise reproduce, any relevant licensing terms or ToU, as well as any fees associated with these restrictions.

The dataset is licensed under the license of CC BY-NC-SA.

\item \textbf{Have any third parties imposed IP-based or other restrictions on the data associated with the instances?} If so, please describe these restrictions, and provide a link or other access point to, or otherwise reproduce, any relevant licensing terms, as well as any fees associated with these restrictions.

To the best of our knowledge, no third parties have imposed any IP-based or other restrictions on the data associated with the instances.

\item \textbf{Do any export controls or other regulatory restrictions apply to the dataset or to individual instances?} If so, please describe these restrictions, and provide a link or other access point to, or otherwise reproduce, any supporting documentation.

Based on our knowledge, there are no export controls or other regulatory restrictions that apply to this dataset or to individual instances.

\item \textbf{Any other comments?}

None.

\end{itemize}

\subsection{Maintenance}

\begin{itemize}

\item \textbf{Who will be supporting/hosting/maintaining the dataset?}

The authors of this paper.

\item \textbf{How can the owner/curator/manager of the dataset be contacted (e.g., email address)?}

We will provide our email address.

\item \textbf{Is there an erratum?} If so, please provide a link or other access point.

No.

\item \textbf{Will the dataset be updated (e.g., to correct labeling
    errors, add new instances, delete instances)?} If so, please
  describe how often, by whom, and how updates will be communicated to
  dataset consumers (e.g., mailing list, GitHub)?

Yes. However, the frequency is not determined. We will publish the updated dataset on the same repository and announce it.

\item \textbf{Will older versions of the dataset continue to be
    supported/hosted/maintained?} If so, please describe how. If not,
  please describe how its obsolescence will be communicated to dataset
  consumers.

Yes. The older versions of the dataset will still be available on the same repository.

\item \textbf{If others want to extend/augment/build on/contribute to
    the dataset, is there a mechanism for them to do so?} If so,
  please provide a description. Will these contributions be
  validated/verified? If so, please describe how. If not, why not? Is
  there a process for communicating/distributing these contributions
  to dataset consumers? If so, please provide a description.

Yes. Through emails.

\item \textbf{Any other comments?}

None.

\end{itemize}

\end{document}